\pdfoutput=1
\documentclass[11pt]{article}
\PassOptionsToPackage{table}{xcolor}
\usepackage[preprint]{acl}
\usepackage{times,latexsym}
\usepackage[T1]{fontenc}
\usepackage[utf8]{inputenc}
\usepackage{microtype,inconsolata}
\usepackage{graphicx,booktabs,tabularx,amsmath,amssymb}
\usepackage{placeins}
\usepackage{fontawesome5}
\newcounter{algorithm}
\title{\raisebox{-0.32em}{\includegraphics[height=1.35em]{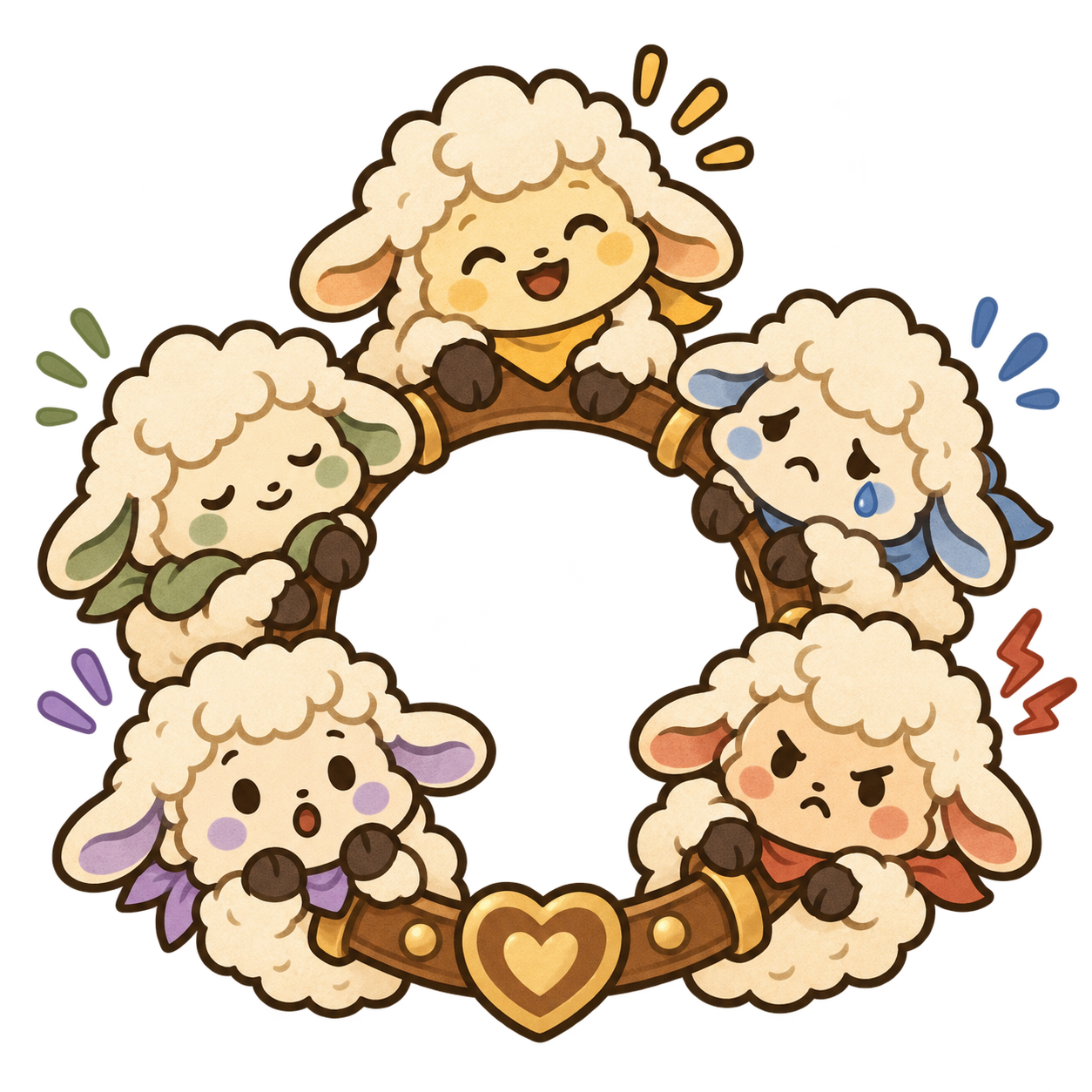}}\hspace{0.25em}\textcolor{mheAccent}{Emo}\textcolor{studentBlue}{Harness}: Boosting Multimodal \\ Emotion Understanding with Meta-Harness Evolving}
\author{  
\textbf{Fenghai Li\textsuperscript{1}},
\textbf{Bingxuan Li\textsuperscript{1}},
\textbf{Jiawen Zhang\textsuperscript{2}},
\textbf{Yifan Shen\textsuperscript{1}},
\\
\textbf{Jiateng Liu\textsuperscript{1}},
\textbf{Yutong Zhang\textsuperscript{3}},
\textbf{Yue Guo\textsuperscript{1}}
\\[0.4em]
\normalsize
\textsuperscript{1}University of Illinois Urbana-Champaign\quad
\textsuperscript{2}University of California, Davis\quad
\textsuperscript{3}Stanford University
\\[0.35em]
\faGlobe\ \href{https://emo-harness.github.io}{https://emo-harness.github.io}
}
\usepackage{xcolor}
\definecolor{mheAccent}{HTML}{6A3D9A}
\definecolor{mheTint}{HTML}{F1ECF7}
\definecolor{studentBlue}{HTML}{3F6EA5}
\definecolor{audioAmber}{HTML}{926316}
\definecolor{inkGray}{HTML}{536474}
\definecolor{errRed}{HTML}{B5473A}
\newcommand{\mhe}{Meta-Harness Evolving}
\newcommand{\mheshort}{MHE}
\newcommand{\EmoHarness}{\texorpdfstring{\textcolor{mheAccent}{Emo}\textcolor{studentBlue}{Harness}}{EmoHarness}}
\definecolor{hdrInk}{HTML}{3B4252}
\usepackage{multirow}


\NewDocumentCommand{\jiateng}
{ mO{} }{\textcolor{blue}{\textsuperscript{\textit{jiateng}}\textsf{\textbf{\small[#1]}}}}

\begin{document}
\raggedbottom
\maketitle
{\renewcommand\thefootnote{}%
\footnotetext{Correspondence to: \{max7,bl61,yueg\}@illinois.edu.}}

\begin{abstract}
Small multimodal models can misjudge emotion when words, voice, and visual cues conflict. We present \EmoHarness{}, which runs such a model, the frozen \emph{student}, inside a \emph{harness}: the prompts, settings, and verification code around its calls. A human-written starting harness, the \emph{scaffold}, numbers the student's visual observations, transcript, and prosody descriptions as citable evidence and has the student weigh candidate emotions against it. In \emph{\mhe{}} (\mheshort{}), a stronger LLM, the \emph{Builder}, reads the student's errors and execution traces offline and edits the harness; static checks and validation gates decide which edits are kept, and the Builder never sees evaluation items. On 5,507 MME-Emotion items, \EmoHarness{} raises task-macro Recognition by 7.6--13.4 points over a single video-only call for all four students and both Builders. Transcripts and prosody descriptions supply most of this boost ($+5.38$ to $+10.71$); the scaffold changes the audio-augmented call by $-0.79$ to $+2.54$, partly through extra multi-label predictions, and evolved edits add at most $+0.95$, although no evolved increment survives Holm correction. The harness's emotion context also transfers to IntentQA, with small gains for three of four students, none of which survives correction. \EmoHarness{} thus boosts small multimodal students chiefly by supplying and organizing audio-derived evidence.
\end{abstract}

\begin{figure}[t]
\centering
\includegraphics[width=\columnwidth]{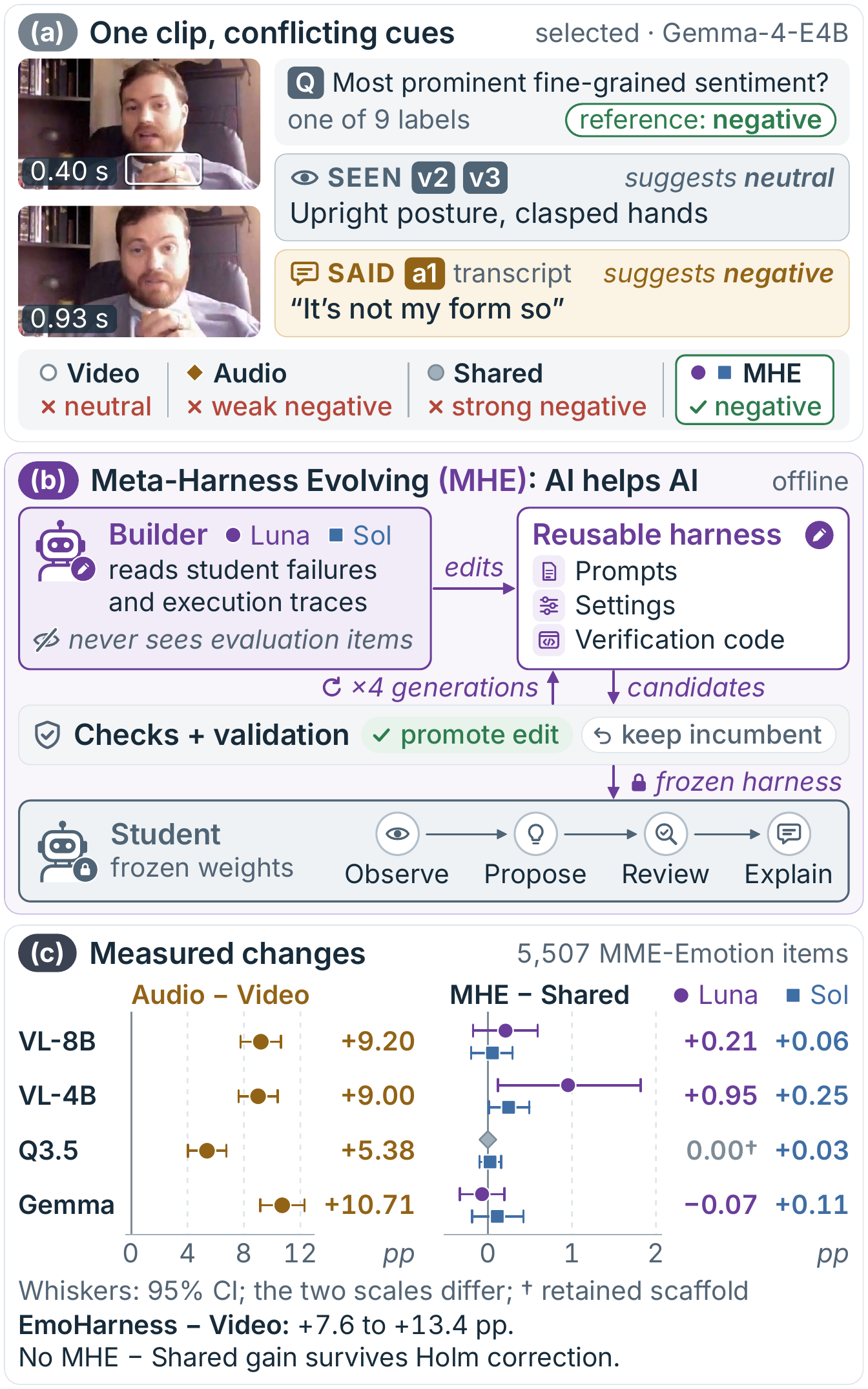}
\caption{\textbf{\EmoHarness{} at a glance.} \textbf{(a)}~A selected held-out clip (Gemma-4-E4B; Figure~\ref{fig:case}a): posture suggests \emph{neutral}, the words \emph{negative}. Video, Audio (video plus transcript and prosody text), and Shared (the human-written scaffold) miss the reference; both \mheshort{} harnesses answer \emph{negative}. \textbf{(b)}~Offline, a Builder LLM (GPT-5.6 Luna or GPT-6.1 Sol) edits the harness from the frozen student's failures and execution traces; checks and validation gates promote edits over four generations. \textbf{(c)}~Task-macro Recognition contrasts (VL: Qwen3-VL; Q3.5: Qwen3.5-4B).}
\vspace{-0.2in}
\label{fig:teaser}
\end{figure}

\section{Introduction}
When someone says ``That's great,'' a pause or a tense expression may reflect sarcasm rather than satisfaction. Understanding emotion requires interpreting words, voice, and visual behavior together, particularly when these cues conflict \citep{zadeh2018mosei,shen2026cogniroute}. 
Misreading them can have consequences beyond an incorrect label: a system may misunderstand a person's intent, respond inappropriately to a social interaction, or overlook a patient's indirectly expressed fear or discomfort \citep{li2023intentqa,wilf2023socialiq,han2026medconceal}.
For smaller multimodal models, transcripts and prosody descriptions supplement visual observations with evidence about what is said and how it is delivered \citep{cao2024visual,li2026toward,li2026trimming}. But interpreting these sources together remains challenging, especially when they suggest different emotions. We therefore ask whether another AI can help a small multimodal student use this evidence more effectively without updating the student's weights. 

Existing emotion models adapt weights through instruction tuning or reinforcement learning \citep{cheng2024emotionllama,zhao2025r1omni}. An alternative is to adapt frozen systems through model-generated instructions and program optimization \citep{zhou2023ape,opsahlong2024mipro,hu2025adas}. 
We build on this approach but direct adaptation toward 
a student's recurring emotion-label failures and cross-modal conflicts.

We introduce \textbf{\EmoHarness{}} (Figure~\ref{fig:teaser}), an AI-for-AI (AI4AI) framework that adapts emotion inference through \emph{\mhe{}} (MHE). An offline Builder examines emotion-label confusions, cross-modal conflicts, and execution traces, and then proposes reusable edits to prompts, execution settings, and restricted verification code. The frozen student organizes visual observations, transcripts, and prosody descriptions into indexed textual evidence, proposes alternative emotional interpretations, and reviews their support and contradictions. Checks and validation select a harness that is frozen before evaluation. If no edit passes, the human-written scaffold is retained. The Builder does not answer evaluation questions.

To assess the contribution of Builder-guided adaptation, we compare the evolved harness (\emph{MHE}) with its initial human-written version (\emph{Shared}) and with single-call prediction from video supplemented with transcripts and prosody descriptions (\emph{Audio}). Comparing Audio with single-call video-only prediction (\emph{Video}) measures the effect of adding this audio-derived evidence. We also examine transfer to task types excluded from harness construction and test whether emotion context generated without access to downstream questions helps intent question answering.

Across four students and two Builders, \EmoHarness{} improves task-macro Recognition on 5,507 MME-Emotion items by 7.6--13.4 points over Video. Most of this improvement comes from transcripts and prosody descriptions: Audio outperforms Video by 5.38--10.71 points, whereas Builder-guided adaptation produces much smaller changes relative to Shared ($-0.07$ to $+0.95$ points). Adding the harness's emotion context also yields small observed gains for three of four students on IntentQA. 

In summary, we introduce \EmoHarness{}, a failure-driven harness construction framework, which boosts frozen multimodal students on emotion recognition and transfers its emotion context to intent question answering. In addition, we conduct a systematic analysis of where the boost comes from and what limits harness evolution. We believe our work paves the way for more adaptive and generalizable multimodal systems that can systematically learn from failures, improve without updating model parameters, and transfer acquired capabilities across related tasks.

\section{Related Work}
\paragraph{Multimodal emotion understanding.}
Emotion-LLaMA and R1-Omni adapt model weights through instruction tuning and reinforcement learning, respectively \citep{cheng2024emotionllama,zhao2025r1omni}; \EmoHarness{} instead adapts the inference program around frozen weights. We evaluate on MME-Emotion \citep{zhang2025mme} and reuse emotion context on IntentQA \citep{li2023intentqa}. Following the multi-dimensional evaluation of HEMM \citep{liang2024hemm}, we separate recognizing an emotion from producing useful downstream context.

\paragraph{Optimizing prompts, programs, and harnesses.}
APE, OPRO, and ProTeGi search instructions with model-generated proposals or feedback \citep{zhou2023ape,yang2024opro,pryzant2023protegi}; DSPy and MIPRO optimize multi-stage programs \citep{khattab2024dspy,opsahlong2024mipro}; TextGrad and Trace learn from textual feedback and execution traces \citep{yuksekgonul2025textgrad,cheng2024trace,xu2026steve}; and ADAS and AFlow search code-defined agents \citep{hu2025adas,zhang2025aflow}. Meta-Harness exposes prior code, scores, and traces to a proposer \citep{lee2026metaharness}, automated harness adaptation can target the failures of smaller models \citep{yang2026better}, and multi-agent and self-evolving systems learn from feedback \citep{li2025metal,li-etal-2026-pearl,li2026echofoley}. \mheshort{} applies this line of work to multimodal emotion recognition: a student's emotion confusions and cross-modal conflicts drive reusable edits, a separate validation split gates every promotion, and the evolved harness is frozen and reused, unlike programs generated for each request \citep{gupta2023visprog}.

\paragraph{Feedback and procedural adaptation.}
Self-Refine revises single outputs and Reflexion carries feedback across trials \citep{madaan2023selfrefine,shinn2023reflexion}; \EmoHarness{} separates the student's per-instance review from the Builder's offline edits. Self-correction can fail \citep{huang2024selfcorrect}, agent errors can cascade across steps \citep{zhu2025llm}, and explanations need not reflect decisions \citep{turpin2023unfaithful}. We therefore score harnesses only on items the Builder never saw, report validation and evaluation gains separately, and count student calls \citep{kapoor2025agents}.

\section{\EmoHarness{}: \mhe{} for Emotion Understanding}
\label{sec:inference}
\begin{figure*}[!t]
\centering
\includegraphics[width=\textwidth]{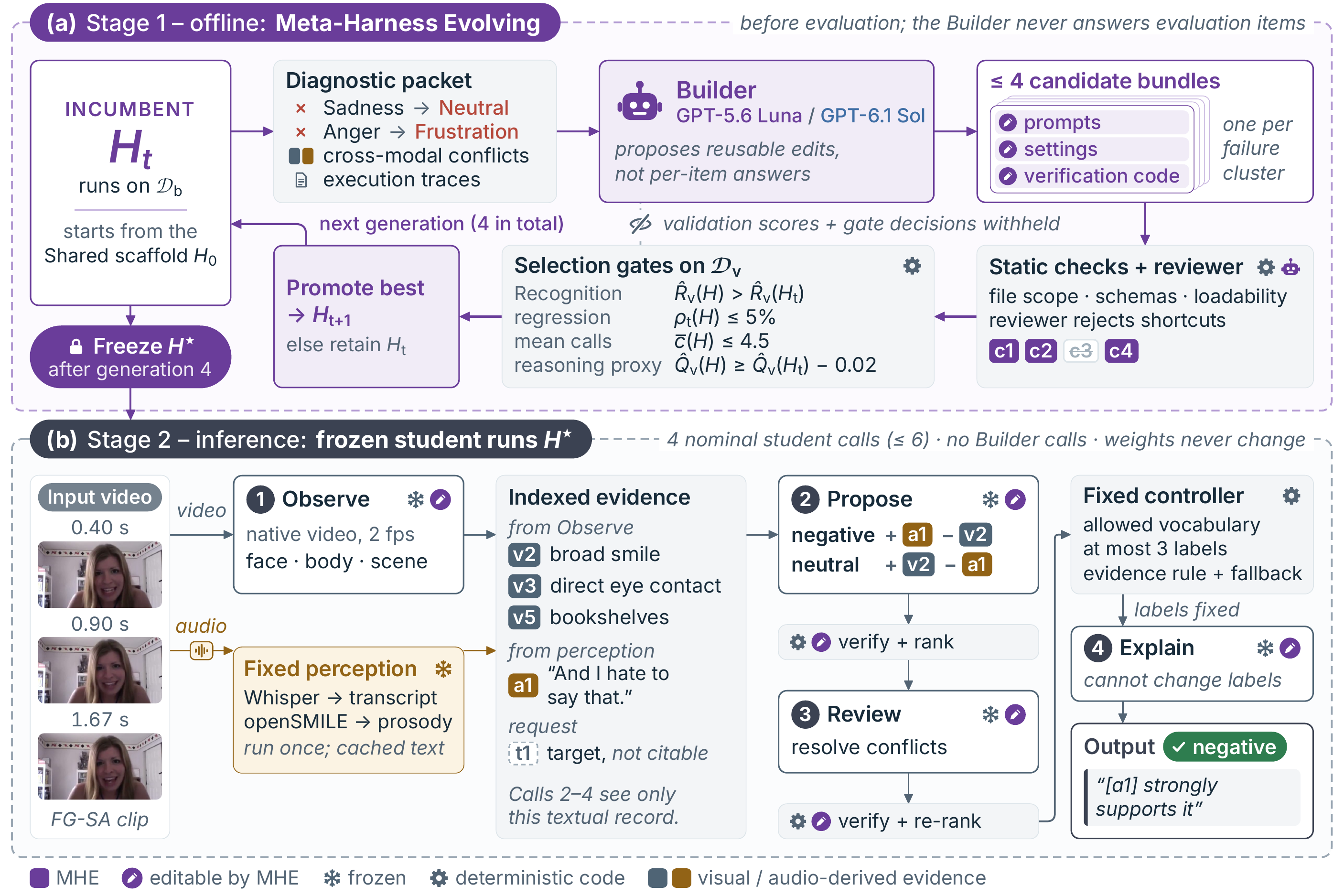}
\caption{\textbf{How \mhe{} builds a harness and how the frozen student runs it.} \textbf{(a)}~Offline, the Builder turns the incumbent's construction failures and traces into at most four candidate edits per generation; static checks, a proposal reviewer, and the gates of Eq.~\ref{eq:gates} promote the best candidate or retain the incumbent, and after four generations the harness is frozen. \textbf{(b)}~At inference the frozen student makes four calls (at most six with retries) around deterministic verification and a fixed controller. The example is the fine-grained sentiment (FG-SA) clip of Appendix Figure~\ref{fig:case_appendix}a: the evidence and the quoted output come from the GPT-5.6 Luna harness on Qwen3-VL-4B; the Propose rows, the confusions in (a), and c1--c4 are illustrative.}
\label{fig:framework}
\end{figure*}

A harness is a \emph{bundle} of prompts, settings files, and verification code. Figure~\ref{fig:framework} shows how \mheshort{} evolves one offline and how the student runs it.

\subsection{Task, Roles, and Fixed Boundaries}
Given a video $v$, a request $q$, and allowed labels $\mathcal{Y}_q$, a harness $H$ invokes the frozen student $M_\theta$ and returns labels $\hat Y$, an explanation $z$, and a trace $\tau$:
\begin{equation}
(\hat Y,z,\tau)=H(M_\theta;v,P(v),q,\mathcal{Y}_q).
\end{equation}
Fixed preprocessing $P(v)$ supplies a transcript, prosodic descriptors, and a frame cache. \mheshort{} changes $H$ but never $\theta$, and within $H$ the perception tools, evidence and output schemas, and output controller are also fixed; the Builder may rewrite prompts, settings (e.g., transcript presentation), and restricted verification code.

\subsection{Evidence-Indexed Emotion Inference}
Audio preprocessing uses Whisper large-v3 (via faster-whisper) and openSMILE eGeMAPSv02 \citep{radford2023whisper,eyben2010opensmile,eyben2016gemaps}; reference-normalized acoustic features become descriptions such as \emph{high pitch}, never emotion labels. The human-written scaffold $H_0$ (\emph{Shared}) uses four student calls. \emph{Observe} extracts facial, bodily, scene, and temporal cues from the video, decoded at 2~fps; deterministic assembly indexes them ($v_1,v_2,\ldots$) along with transcript sentences and prosody phrases ($a_1,a_2,\ldots$), and later calls see only this textual record. \emph{Propose} lists candidate labels with confidences and with identifiers of supporting and contradicting evidence; deterministic code checks the identifiers against the record and then ranks the candidates. \emph{Review} re-examines the ranked candidates, and its output is verified the same way. A fixed controller then selects at most three labels from the vocabulary. It prefers candidates whose valid support exceeds their contradiction and fills any remaining slots from the ranking, so it always returns an answer, even one without valid support. \emph{Explain} then writes an explanation that cannot change the labels. Each malformed call may be retried once, under a cap of six calls (Appendix~\ref{app:method}).

\subsection{\mhe{}}
\label{sec:builder}
\mheshort{} evolves the harness over $T=4$ generations. In each, the Builder proposes candidate edits, and gates either \emph{promote} one candidate or keep the current harness. \mheshort{} inherits from Meta-Harness \citep{lee2026metaharness} a proposer that reads prior harness code and execution traces. It differs in three respects. First, the Builder sees the failures of one frozen student, clustered by gold--prediction confusion and by cross-modal conflict; a deterministic check flags a conflict when the student's visual and audio-derived citations favor different labels. Second, gates on a separate validation split decide promotion, and the Builder sees neither validation scores nor gate decisions, only the resulting harness. Third, each run keeps one current harness, the \emph{incumbent}, giving a chain $H_0\rightarrow\cdots\rightarrow H_T$.

\paragraph{Variation.}
The incumbent runs on a construction split $\mathcal{D}_b$, whose failures the Builder may read. Given the incumbent's files, intermediate records, clustered failures, and earlier proposals with check and reviewer feedback, the Builder proposes up to four candidate harnesses, each an edit aimed at one failure cluster; evaluation data are never exposed.

\paragraph{Selection.}
Candidates must pass file-scope, schema, loadability, and static-code checks, as well as a proposal reviewer that rejects, for example, edits that force a label regardless of evidence or add unsupported settings; these are filters, not guarantees. Survivors run on the validation split $\mathcal{D}_v$. Let $r_i(H)\in[0,1]$ be per-item Recognition (Section~\ref{sec:metrics}) and $\mathcal{S}_t$ the validation items solved ($r_i=1$) by any incumbent so far. The regression rate is the fraction of $\mathcal{S}_t$ that a candidate no longer solves:
\begin{equation}
\rho_t(H)=
\frac{|\mathcal{S}_t\setminus\operatorname{Solved}(H;\mathcal{D}_v)|}
{|\mathcal{S}_t|}.
\label{eq:regression}
\end{equation}
A candidate $H$ replaces the incumbent $H_t$ only if
\begin{equation}
\begin{gathered}
\widehat R_v(H)>\widehat R_v(H_t),\qquad
\rho_t(H)\leq\delta,\\
\overline c(H)\leq B,\qquad
\widehat Q_v(H)\geq\widehat Q_v(H_t)-\epsilon,
\end{gathered}
\label{eq:gates}
\end{equation}
where $\widehat R_v$ is the mean validation Recognition, $\overline c$ the mean number of student calls, and $\widehat Q_v$ a reasoning proxy used only for selection: the average share of non-final reasoning steps that a judge marks correct, on a fixed 200-item validation subset (Appendix~\ref{app:method}). We use $\delta=0.05$, $B=4.5$, and $\epsilon=0.02$. The best passing candidate, ranked by validation Recognition and then by fewer calls, becomes $H_{t+1}$; otherwise $H_{t+1}=H_t$. After $T$ generations the evolved harness $H^\star=H_T$ is frozen; if nothing was promoted, $H^\star$ is the Shared scaffold itself. Algorithm~\ref{alg:builder} (Appendix~\ref{app:method}) summarizes the loop.

\section{Experimental Setup}
\label{sec:setup}
We ask what contributes to recognition (RQ1), whether evolved edits transfer to held-out task types (RQ2), and whether the harness's emotion context helps intent question answering (RQ3).

\subsection{Students, Builders, and Data}
\paragraph{Models.}
We serve Qwen3-VL-8B, Qwen3-VL-4B \citep{bai2025qwen3vl}, Qwen3.5-4B \citep{qwen2026qwen35}, and Gemma-4-E4B \citep{gemma2026gemma4} with vLLM \citep{kwon2023vllm} at temperature zero, with thinking disabled for Qwen3.5 and Gemma. Audio reaches every student only as transcript and prosody text. Each student's harness is evolved separately by GPT-5.6 Luna and by GPT-6.1 Sol; each Builder also acts as its own proposal reviewer, and GPT-5.6 Luna scores the reasoning proxy. Each student--Builder pair has one four-generation run. A third Builder configuration, which changes no conclusion, is described in Appendix~\ref{app:implementation}.

\paragraph{Data.}
Construction and validation use 399 and 394 MME-Emotion items \citep{zhang2025mme} from three task types (ER-Lab, ER-Wild: lab and in-the-wild emotion recognition; SA: sentiment analysis). Evaluation uses 5,507 further items, namely the benchmark minus every item whose media overlap construction, validation, or a reserved 129-item set. The evaluation set contains 2,963 items in the seen types and 2,544 in five held-out types (noisy, fine-grained, and multi-label emotion recognition: Noise-ER, FG-ER, ML-ER; fine-grained sentiment: FG-SA; intent recognition: IR). \emph{Held-out} refers to task type, not to an untouched test set: 986 evaluation items, which we cannot identify, were inspected during development. Inspection could have influenced any human design choice, most plausibly the scaffold, but not the Builder's edits directly: the Builder never saw evaluation items (Appendix~\ref{app:provenance}).

\subsection{Conditions and Metrics}
\label{sec:metrics}
\emph{Video} prompts the student once with the video, and \emph{Audio} adds transcript and prosody text to that call. \emph{Shared} runs the four-call scaffold $H_0$, and \emph{\mheshort{}} runs the frozen evolved harness $H^\star$. Recognition, computed locally without an LLM judge, scores the first normalized prediction for single-label requests and the fraction of gold labels recovered for multi-label requests, a recall-like score that ignores extra labels. Audio$-$Video measures added evidence, and \mheshort{}$-$Shared measures what evolution adds. Shared$-$Audio mixes four changes: a multi-step procedure, four calls instead of one, later calls that see only a text record of the video, and more predicted labels, which the recall-like score rewards. Task-macro Recognition averages the eight task means:
\begin{equation}
R_{\mathrm{macro}}=\frac{1}{K}\sum_{k=1}^{K}
\frac{1}{|\mathcal{D}_k|}\sum_{i\in\mathcal{D}_k}r_i .
\label{eq:macro}
\end{equation}
Differences are in percentage points (pp). Intervals are pointwise 95\% task-stratified paired bootstrap intervals (10,000 resamples), and $p$-values are two-sided bootstrap values. The eight \mheshort{}$-$Shared tests form one Holm family \citep{holm1979simple}.

\subsection{Downstream Reuse of Emotion Context}
\label{sec:downstream}
We use the IntentQA test split (2,134 questions on 567 videos). Each evolved harness writes one emotion context per video (predicted emotion, alternatives, and a four-step analysis) using the ER-Wild instruction and label set, without seeing any question. Without a harness, the student answers each question given the cached frames (\emph{Frames}), the frames with transcript and prosody (\emph{Audio}), or the Audio inputs plus the emotion context (\emph{Audio+Emotion}); answers are capped at 64 tokens (1,024 with a final-answer parser for Qwen3.5-4B), and unparsed replies count as wrong. Intervals use 10,000 paired bootstrap resamples clustered by video, with Holm correction over eight tests.

\section{Results}
\label{sec:results}
\begin{table*}[t]
\centering
\footnotesize
\setlength{\tabcolsep}{2.5pt}
\renewcommand{\arraystretch}{1.0}
\begin{tabularx}{\textwidth}{@{}cl*{11}{>{\centering\arraybackslash}X}>{\hspace{5pt}}c}
\toprule
 & & \multicolumn{3}{c}{\textit{Seen task types}} & \multicolumn{5}{c}{\textit{Held-out task types}} & \multicolumn{3}{c}{\textit{Summary}} & \\
\cmidrule(lr){3-5}\cmidrule(lr){6-10}\cmidrule(lr){11-13}
 &  & ER- & ER- &  & Noise- & FG- & ML- & FG- &  &  & Held- &  &  \\
 &  & Lab & Wild & SA & ER & ER & ER & SA & IR & Seen & out & Macro & 95\% CI \\
\midrule
 & Video & 8.5 & 36.2 & 41.4 & 56.4 & 23.6 & 38.8 & 13.2 & 21.5 & 28.7 & 30.7 & 29.95 & \textcolor{inkGray}{\scriptsize 28.8--31.1} \\
 & Audio & 23.2 & \textbf{38.0} & \textbf{63.3} & \textbf{65.0} & 19.2 & 41.7 & \textbf{30.8} & 32.0 & 41.5 & 37.7 & 39.15 & \textcolor{inkGray}{\scriptsize 37.8--40.5} \\
 & Shared & 28.8 & 35.5 & 59.9 & 56.6 & 26.1 & 48.9 & 26.7 & 41.2 & 41.4 & 39.9 & 40.47 & \textcolor{inkGray}{\scriptsize 39.0--41.9} \\
\rowcolor{mheTint}\cellcolor{white} & \mbox{\textbf{\mheshort{}}~{\scriptsize GPT-5.6~Luna}} & \textbf{29.4} & 35.0 & 60.6 & 57.2 & 26.2 & 48.7 & 27.8 & 40.6 & \textbf{41.7} & \textbf{40.1} & \textbf{40.68} & \textcolor{inkGray}{\scriptsize 39.2--42.1} \\
\rowcolor{mheTint}\cellcolor{white}\multirow{-5}{*}{\rotatebox[origin=c]{90}{\scriptsize Qwen3-VL-8B}} & \mbox{\textbf{\mheshort{}}~{\scriptsize GPT-6.1~Sol}} & 28.2 & 35.7 & 60.5 & 56.4 & \textbf{26.4} & \textbf{49.1} & 26.7 & 41.2 & 41.5 & 40.0 & 40.52 & \textcolor{inkGray}{\scriptsize 39.1--41.9} \\
\midrule
 & Video & 8.5 & 34.7 & 42.6 & 51.8 & \textbf{21.9} & 36.9 & 11.1 & 27.6 & 28.6 & 29.9 & 29.39 & \textcolor{inkGray}{\scriptsize 28.2--30.6} \\
 & Audio & 18.6 & \textbf{37.0} & 63.8 & \textbf{61.4} & 20.8 & 39.4 & \textbf{35.1} & 31.0 & 39.8 & \textbf{37.5} & 38.39 & \textcolor{inkGray}{\scriptsize 37.0--39.7} \\
 & Shared & 22.6 & 35.6 & 65.6 & 57.4 & 17.7 & 45.4 & 20.8 & 35.7 & 41.3 & 35.4 & 37.60 & \textcolor{inkGray}{\scriptsize 36.2--39.0} \\
\rowcolor{mheTint}\cellcolor{white} & \mbox{\textbf{\mheshort{}}~{\scriptsize GPT-5.6~Luna}} & \textbf{23.7} & 35.7 & 65.7 & 58.6 & 18.4 & \textbf{46.4} & 21.9 & \textbf{37.9} & \textbf{41.7} & 36.7 & \textbf{38.56} & \textcolor{inkGray}{\scriptsize 37.2--39.9} \\
\rowcolor{mheTint}\cellcolor{white}\multirow{-5}{*}{\rotatebox[origin=c]{90}{\scriptsize Qwen3-VL-4B}} & \mbox{\textbf{\mheshort{}}~{\scriptsize GPT-6.1~Sol}} & 23.2 & 35.9 & \textbf{65.8} & 57.8 & 18.1 & 45.5 & 20.8 & 35.7 & 41.6 & 35.6 & 37.85 & \textcolor{inkGray}{\scriptsize 36.5--39.2} \\
\midrule
 & Video & 14.1 & 38.1 & 38.3 & 46.8 & 20.7 & \textbf{46.7} & 6.1 & 26.2 & 30.2 & 29.3 & 29.62 & \textcolor{inkGray}{\scriptsize 28.4--30.8} \\
 & Audio & 20.3 & \textbf{38.2} & 56.1 & \textbf{51.2} & 20.8 & 39.2 & 21.7 & 32.5 & 38.2 & 33.1 & 34.99 & \textcolor{inkGray}{\scriptsize 33.7--36.4} \\
 & Shared & 29.9 & 33.4 & 60.6 & 45.6 & 23.8 & 45.7 & 24.9 & 33.9 & 41.3 & 34.8 & 37.23 & \textcolor{inkGray}{\scriptsize 35.8--38.6} \\
\rowcolor{mheTint}\cellcolor{white} & \mbox{\textbf{\mheshort{}}~{\scriptsize GPT-5.6~Luna}}\rlap{$^{\dagger}$} & 29.9 & 33.4 & 60.6 & 45.6 & 23.8 & 45.7 & 24.9 & 33.9 & 41.3 & 34.8 & 37.23 & \textcolor{inkGray}{\scriptsize 35.8--38.6} \\
\rowcolor{mheTint}\cellcolor{white}\multirow{-5}{*}{\rotatebox[origin=c]{90}{\scriptsize Qwen3.5-4B}} & \mbox{\textbf{\mheshort{}}~{\scriptsize GPT-6.1~Sol}} & 29.9 & 33.6 & 60.6 & 45.6 & 23.8 & 45.7 & 24.9 & 33.9 & \textbf{41.4} & 34.8 & \textbf{37.25} & \textcolor{inkGray}{\scriptsize 35.8--38.6} \\
\midrule
 & Video & 9.0 & 30.8 & 39.8 & 41.0 & 25.1 & 36.0 & 9.1 & 25.4 & 26.6 & 27.3 & 27.03 & \textcolor{inkGray}{\scriptsize 25.8--28.2} \\
 & Audio & 27.1 & \textbf{33.4} & 61.2 & \textbf{52.4} & 20.2 & 40.4 & 32.8 & 34.5 & 40.6 & 36.1 & 37.74 & \textcolor{inkGray}{\scriptsize 36.3--39.2} \\
 & Shared & 29.4 & 31.0 & \textbf{64.9} & 47.4 & 26.0 & 47.2 & 38.8 & 37.5 & 41.7 & 39.4 & 40.28 & \textcolor{inkGray}{\scriptsize 38.8--41.7} \\
\rowcolor{mheTint}\cellcolor{white} & \mbox{\textbf{\mheshort{}}~{\scriptsize GPT-5.6~Luna}} & 29.4 & 31.5 & 64.5 & 47.2 & 26.0 & 47.6 & 38.6 & 36.9 & \textbf{41.8} & 39.3 & 40.21 & \textcolor{inkGray}{\scriptsize 38.8--41.7} \\
\rowcolor{mheTint}\cellcolor{white}\multirow{-5}{*}{\rotatebox[origin=c]{90}{\scriptsize Gemma-4-E4B}} & \mbox{\textbf{\mheshort{}}~{\scriptsize GPT-6.1~Sol}} & 29.4 & 31.1 & 64.2 & 46.6 & 25.9 & \textbf{48.5} & \textbf{39.7} & \textbf{37.7} & 41.6 & \textbf{39.7} & \textbf{40.39} & \textcolor{inkGray}{\scriptsize 38.9--41.9} \\
\bottomrule
\end{tabularx}
\captionsetup{position=bottom}
\caption{\textbf{Main results on MME-Emotion.} Recognition (\%) per task type and task-macro averages (Seen, Held-out, Macro) on 5{,}507 items, for each student under Video and Audio (one call each), the Shared scaffold, and the harnesses evolved by \mhe{} (\mheshort{}) with each Builder (about four calls each). Seen task types occur in construction; held-out types do not. FG-ER and ML-ER use recall-like Recognition. Shaded rows: \mheshort{}; \textbf{bold}: best per student and column (ties unmarked). 95\% CI: task-stratified bootstrap interval of the macro score (10{,}000 resamples). $^{\dagger}$\,No candidate passed the gates, so the harness equals Shared. No \mheshort{}$-$Shared difference survives Holm correction over eight tests (smallest adjusted $p=.192$). Differences reported in the text and figures use unrounded scores and can differ by 0.01 (0.1 in one-decimal columns) from differences of the displayed values.}
\label{tab:main}
\end{table*}

\begin{figure}[t]
\centering
\includegraphics[width=\columnwidth]{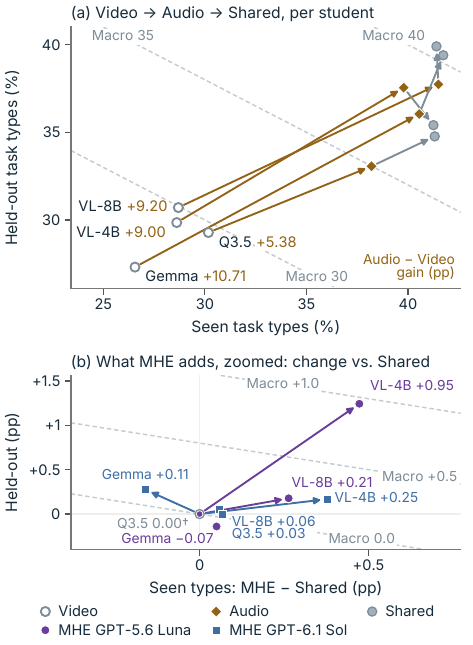}
\caption{\textbf{Recognition paths per student.} \textbf{(a)}~Seen- versus held-out-type macro Recognition (\%; Table~\ref{tab:main}): arrows run Video (open circle) $\rightarrow$ Audio (amber diamond) $\rightarrow$ Shared (gray); amber numbers beside each name: Audio$-$Video macro gain (pp); dashed lines: equal overall Macro. \textbf{(b)}~\mheshort{}$-$Shared, zoomed (origin = Shared; ticks 10$\times$ finer than in (a)); violet circle: GPT-5.6 Luna; blue square: GPT-6.1 Sol; changes below 0.1 points have no arrow; labels and dashed lines give the macro change. $^{\dagger}$\,Harness equals Shared. Short names and rounding as in Table~\ref{tab:context}.}
\label{fig:recognition_paths}
\end{figure}

\subsection{RQ1: Where Do the Gains Come From?}
\label{sec:rq1}
\paragraph{Added evidence dominates.}
Transcript and prosody text give every student its largest gain (Table~\ref{tab:main}, Figure~\ref{fig:recognition_paths}a; Appendix Table~\ref{tab:builder_inference}b): Audio$-$Video ranges from $+5.38$ (Qwen3.5-4B) to $+10.71$ points (Gemma-4-E4B), and every interval excludes zero. Shared$-$Audio is $+2.23$ (Qwen3.5-4B) and $+2.54$ (Gemma-4-E4B), whose intervals exclude zero, and $+1.32$ (Qwen3-VL-8B) and $-0.79$ (Qwen3-VL-4B), whose intervals include zero. The Qwen3.5-4B and Gemma-4-E4B gains depend partly on recall-like scoring of FG-ER and ML-ER, where Shared predicts more labels than Audio. Scoring these two tasks by set F1 lowers Shared$-$Audio to $+1.10$ (Qwen3.5-4B), $+1.00$ (Gemma-4-E4B), $-0.25$ (Qwen3-VL-8B), and $-1.41$ (Qwen3-VL-4B); under F1, every Qwen3-VL harness, scaffold or evolved, scores below single-call Audio (Appendix~\ref{app:recognition}). Evolved harnesses exceed Video by 7.61--13.36 points and differ from single-call Audio by $-0.54$ to $+2.65$.

\paragraph{What \mhe{} adds.}
Across the eight student--Builder pairs, \mheshort{}$-$Shared ranges from $-0.07$ to $+0.95$ points (Figure~\ref{fig:recognition_paths}b). The largest increment, GPT-5.6 Luna on Qwen3-VL-4B ($+0.95$, 95\% CI [$+0.12$, $+1.82$]), does not survive Holm correction (adjusted $p=.192$), and neither does any other contrast (Appendix~\ref{app:recognition}). GPT-6.1 Sol's harnesses score 0.03 to 0.25 points above Shared for every student; only the Qwen3-VL-4B interval excludes zero (lower bound $+0.01$).

\paragraph{Validation gains shrink on evaluation.}
Across four runs per Builder, the gates promote six GPT-5.6 Luna edits and five GPT-6.1 Sol edits, raising validation Recognition by up to 2.54 points (Appendix Figure~\ref{fig:trajectories}). Scored like validation (flat item mean) on the 2,963 seen-type evaluation items, the seven promoted harnesses change by $-0.20$ to $+0.30$ points; the gain of GPT-5.6 Luna's Qwen3-VL-8B harness falls from 2.54 to 0.10 points.

\subsection{RQ2: Transfer to Held-Out Task Types}
With one exception, held-out changes are as small as seen-type changes. \mheshort{}$-$Shared ranges from $-0.16$ to $+0.47$ points on the seen types and, for seven pairs, from $-0.14$ to $+0.28$ on the five held-out types (Table~\ref{tab:main}, Figure~\ref{fig:recognition_paths}b; Appendix Figure~\ref{fig:task_adaptation}). The exception is again GPT-5.6 Luna on Qwen3-VL-4B ($+1.24$), which lifts held-out Recognition to 36.65, still below single-call Audio (37.55). With seen-type gains this small, the held-out results say little about transfer beyond this pair.

\subsection{Case Study: What an Evolved Edit Changes}
\label{sec:case}
\begin{figure*}[t]
\centering
\includegraphics[width=\textwidth]{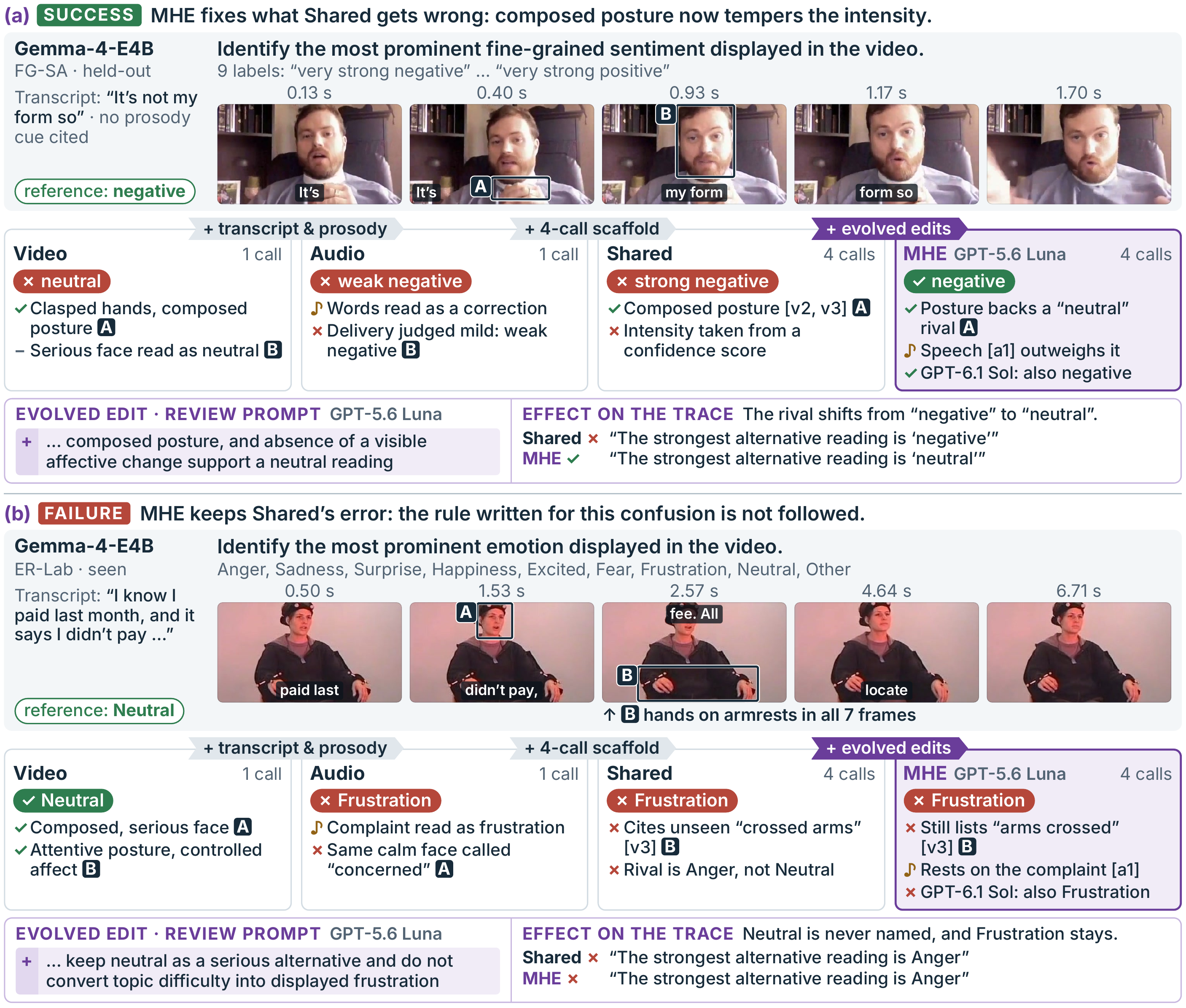}
\caption{\textbf{Case study: \mheshort{} versus baselines on Gemma-4-E4B.} The input is drawn once: instruction, transcript, and five of our seven sampled frames per clip (the student saw 2~fps), subtitled with the words spoken within 0.12\,s of each frame (every condition except Video receives the transcript); lettered boxes mark the regions that cards cite. Each card gives one condition's answer (green: correct; red: wrong), and the tags between cards name what each condition adds. Note glyphs: green check (supported); red cross (misread, unsupported, or wrong); amber note (audio-derived); gray dash (limitation). The last \mheshort{} note gives the answer of the GPT-6.1 Sol harness; [v$i$]/[a$j$]: evidence IDs. The edit panel quotes the rule GPT-5.6 Luna added to the review prompt, with verbatim Shared and \mheshort{} excerpts. Cases are selected, not representative.}
\label{fig:case}
\end{figure*}

Figure~\ref{fig:case} follows two Gemma-4-E4B items. In (a), Shared rates a mildly negative remark as \emph{strong negative} because a confidence score, not an observed cue, ranks that label first. With the GPT-5.6 Luna edit, the review call treats the composed posture as evidence for a neutral reading, and the harness answers the reference \emph{negative} instead. In (b), Video answers the correct \emph{Neutral}, but once the transcript of a billing complaint is added, Audio, Shared, and both \mheshort{} harnesses answer \emph{Frustration}. The GPT-5.6 Luna harness keeps \emph{Frustration} even though its added rule asks it to treat neutral as a serious alternative: its explanation never names \emph{Neutral} and still lists crossed arms that no sampled frame shows. Over all 5,507 items, this harness changes Shared's label set on 169 items: 46 improve, 45 worsen, and 78 score the same (more cases in Appendix~\ref{app:cases}).

\subsection{RQ3: Does Emotion Context Help Intent QA?}
\begin{table}[t]
\centering
\footnotesize
\setlength{\tabcolsep}{1.9pt}
\renewcommand{\arraystretch}{1.1}
\begin{tabularx}{\columnwidth}{@{}l*{4}{>{\centering\arraybackslash}X}@{}}
\toprule
 & VL-8B & VL-4B & Q3.5 & Gemma \\
\midrule
Frames & 70.99 & 67.48 & \textbf{79.90} & 57.31 \\
Audio & 72.63 & 68.42 & 77.84 & 60.59 \\
\rowcolor{mheTint}\mbox{+ Emotion \mbox{\textbf{\mheshort{}}~{\scriptsize Luna}}} & \textbf{73.10} & \textbf{70.29} & 76.34\rlap{$^{\dagger}$} & 61.62 \\
\rowcolor{mheTint}\mbox{+ Emotion \mbox{\textbf{\mheshort{}}~{\scriptsize Sol}}} & 72.87 & 69.92 & 76.43 & \textbf{61.95} \\
\bottomrule
\end{tabularx}
\captionsetup{position=bottom}
\caption{\textbf{Downstream QA accuracy (\%).} Inputs as in Section~\ref{sec:downstream}; 2{,}134 IntentQA test questions. Shaded rows add \mheshort{} emotion context; \textbf{bold}: best per student (ties unmarked). Luna: GPT-5.6~Luna; Sol: GPT-6.1~Sol. VL-8B/VL-4B: Qwen3-VL-8B/4B; Q3.5: Qwen3.5-4B; Gemma: Gemma-4-E4B. No Audio+Emotion$-$Audio contrast survives Holm correction over eight tests (smallest adjusted $p=.085$; Appendix Table~\ref{tab:context_delta}). Differences reported in the text and figures use unrounded scores and can differ by 0.01 from differences of the displayed values. $^{\dagger}$\,Retained scaffold: this context is generated by Shared.}
\label{tab:context}
\end{table}

\begin{figure}[t]
\centering
\includegraphics[width=\columnwidth]{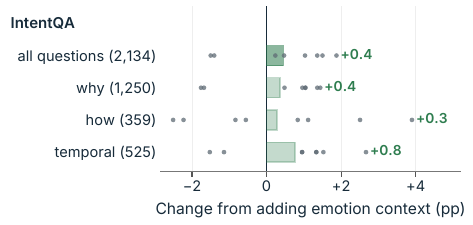}
\caption{\textbf{Emotion context transfers to IntentQA.} Accuracy change (pp) from adding the harness's emotion context to Audio. Bars: mean over 4 students $\times$ 2 Builders (green: gain; red: loss); dots: the 8 individual changes. Rows: all questions, then the official types (temporal: TN+TP). The mean change is positive for every type; none survives Holm correction (Appendix Table~\ref{tab:downstream_strata}).}
\label{fig:downstream_strata}
\end{figure}

On IntentQA, Audio$-$Frames ranges from $-2.06$ to $+3.28$ points (Table~\ref{tab:context}). Adding the emotion context raises accuracy in all six pairs of Qwen3-VL-8B, Qwen3-VL-4B, and Gemma-4-E4B ($+0.23$ to $+1.87$ points; Figure~\ref{fig:downstream_strata}, Appendix Figure~\ref{fig:context}); for Qwen3-VL-4B both pointwise intervals exclude zero ($+1.87$ [$+0.47$, $+3.27$]; $+1.50$ [$+0.05$, $+2.97$]). Qwen3.5-4B loses 1.50 and 1.41 points. Each context changes the correctness of 248 to 306 of the 2,134 answers, so each net change is a small difference between many gains and losses. None of the eight contrasts survives Holm correction (smallest adjusted $p=.085$; Appendix Table~\ref{tab:context_delta}), and the mean change is positive in every official question type, none significantly (Appendix Table~\ref{tab:downstream_strata}). We have no length-matched control (Appendix~\ref{app:pending}), so longer inputs, rather than emotion content, could account for these changes. Scaffold contexts exist only for Qwen3.5-4B, whose GPT-5.6 Luna harness retained the scaffold. Its GPT-6.1 Sol context matches the scaffold's for 531 of 567 videos. No experiment therefore compares a scaffold context with a substantively evolved one, so these changes cannot be attributed to evolution.

\section{Discussion and Conclusion}
\label{sec:discussion}
\paragraph{When does \mheshort{} help?}
The largest gain, GPT-5.6 Luna on Qwen3-VL-4B, occurs for the only student whose scaffold scores below Audio. Its edits add 0.95 points, mostly on held-out types ($+1.24$ versus $+0.47$ on seen types). They change the label set on 1,189 items (21.6\%): 281 improve, 233 worsen, and 675 score the same. Three findings suggest why the other gains are small. First, each promotion picks the best of up to four candidates on 394 items, so the winner's validation gain tends to overstate its gain on new items (Section~\ref{sec:rq1}). The Recognition gate also requires no margin: three of the eleven promotions rest on a single validation item (0.25 points). Second, the regression gate blocked every large validation gain for both Builders: all twelve candidates that gained at least eight validation items (2.03 points) exceeded the 5\% regression budget. Nine were for Qwen3.5-4B, whose prompt edits change the correctness of 56 construction items on average, against 23--27 for the other students. Third, the other evolved harnesses are conservative: each changes the label set on fewer than 4\% of evaluation items. The regression gate protects solved cases at the cost of broad edits; whether a looser gate generalizes better needs a separate test set.

\paragraph{Edits are instructions the student may not follow.}
In Figure~\ref{fig:case}a the added rule moves the trace toward a neutral alternative. In Figure~\ref{fig:case}b the rule written for the Neutral$\rightarrow$Frustration confusion is ignored: Neutral is never named. These selected cases suggest that the student's instruction following, not only the Builder's proposals, can limit what edits achieve. Future gates could check that traces apply edits and that citations match the media.

\paragraph{Emotion context transfers modestly to intent QA.}
The gains for three students are small, and none survives Holm correction (Table~\ref{tab:context}). Qwen3.5-4B, the one student that loses, also loses 2.06 points when only transcript and prosody are added. Its loss may therefore come from added text rather than from emotion content; this student also uses a different answer format (Section~\ref{sec:downstream}). Conditioning the context on the question could strengthen transfer but would reduce reuse across questions.

\paragraph{Conclusion.}
\EmoHarness{} boosts Recognition by 7.6--13.4 points over Video for every student. Most of the boost comes from audio-derived evidence (Audio$-$Video: 5.38--10.71 points). Edits evolved by \mheshort{} add at most 0.95 points over Shared, and none survives Holm correction. The harness's emotion context transfers to IntentQA with small, nonsignificant gains for three of four students.\label{page:main-end}

\section*{Limitations}
\paragraph{Evidence base.}
The media-overlap filter that defines the evaluation split was applied once and has not been independently re-verified. Part of the evaluation set was inspected during development, so held-out results describe unseen task types rather than an untouched test set. The 129 reserved examples remain unevaluated. We score Recognition against gold labels without the benchmark's LLM judge, so our values may differ from published MME-Emotion results, and we do not report its Reasoning and Chain-of-Thought scores.

\paragraph{Construction.}
Each student--Builder pair has one construction trajectory. The two Builders were run about three weeks apart, and the later runs used a slightly extended diagnostic packet. Four generations do not imply matched proposal budgets or compute (Appendix~\ref{app:resources}). The same model acts as Builder and proposal reviewer, and GPT-5.6 Luna also scores the reasoning proxy for both Builders' candidates. The gate values are fixed design choices that we did not vary. Fixed-harness intervals do not capture construction variance, and nonsignificant corrected tests do not establish equivalence. We do not compare against existing optimizers (e.g., MIPRO, TextGrad, or Meta-Harness itself). Prompt-only, evidence-separated, and compute-matched controls are absent, as are Shared-generated contexts except for Qwen3.5-4B, whose GPT-5.6 Luna harness retained the scaffold.

\paragraph{Measurement.}
Visual information can be lost when observations are converted to text. Acoustic descriptions depend on reference statistics, speakers, languages, and recording conditions, and the lack of diarization limits multi-speaker interpretation. MME-Emotion clips include English and Mandarin Chinese speech, and we do not analyze results by language. The three-label cap constrains recall, while recall-like Recognition does not penalize excess labels and favors the scaffold on multi-label tasks (Appendix~\ref{app:recognition}). We do not report Recognition separately on items with cross-modal conflicts, so whether edits help where they are aimed is untested. Valid citations do not establish grounding or faithful explanations, fallback logic can complete unsupported answers, and static checks and model reviewers do not guarantee correct verification code. The 64-token answer cap leaves some downstream replies unparsed, and Qwen3.5-4B instead answers under a 1,024-token cap with a final-answer parser, so answer formatting differs across students.

\section*{Ethical Considerations}
Benchmark emotion labels are interpretations of behavior, not direct measurements of private mental states or clinical conditions. This work does not validate use in diagnosis, employment screening, surveillance, or educational assessment. Cultural, linguistic, and demographic variation requires separate analysis with suitable annotations. Dataset access and redistribution must follow source terms. Videos, transcripts, and construction diagnostics may identify individuals; external model processing and released traces must respect applicable permissions, and the frames reproduced in our case-study figures come from public research datasets, with on-screen names removed. We use MME-Emotion (MIT license) and IntentQA (whose repository states no license) for non-commercial research evaluation only; source videos remain subject to their original terms. The student models are released under Apache 2.0, and Whisper and faster-whisper (MIT), openSMILE (audEERING research license), and vLLM (Apache 2.0) are used under their terms. We release no new data.

\bibliography{custom}

\begin{thebibliography}{43}
\providecommand{\natexlab}[1]{#1}

\bibitem[{Bagher~Zadeh et~al.(2018)Bagher~Zadeh, Liang, Poria, Cambria, and Morency}]{zadeh2018mosei}
AmirAli Bagher~Zadeh, Paul~Pu Liang, Soujanya Poria, Erik Cambria, and Louis-Philippe Morency. 2018.
\newblock \href {https://doi.org/10.18653/v1/P18-1208} {Multimodal language analysis in the wild: {CMU-MOSEI} dataset and interpretable dynamic fusion graph}.
\newblock In \emph{Proceedings of the 56th Annual Meeting of the Association for Computational Linguistics (Volume 1: Long Papers)}, pages 2236--2246. Association for Computational Linguistics.

\bibitem[{Bai et~al.(2025)Bai, Cai, Chen, Chen et~al.}]{bai2025qwen3vl}
Shuai Bai, Yuxuan Cai, Ruizhe Chen, Keqin Chen, et~al. 2025.
\newblock \href {https://arxiv.org/abs/2511.21631} {{Qwen3-VL} technical report}.
\newblock \emph{arXiv preprint arXiv:2511.21631}.

\bibitem[{Cao et~al.(2024)Cao, Shen, Lai, Ye, Ma, Heintz, Chen, Huang, Cao, Zhang et~al.}]{cao2024visual}
Xu~Cao, Yifan Shen, Bolin Lai, Wenqian Ye, Yunsheng Ma, Joerg Heintz, Jintai Chen, Meihuan Huang, Jianguo Cao, Aidong Zhang, et~al. 2024.
\newblock What is the visual cognition gap between humans and multimodal llms?
\newblock \emph{arXiv preprint arXiv:2406.10424}.

\bibitem[{Cheng et~al.(2024{\natexlab{a}})Cheng, Nie, and Swaminathan}]{cheng2024trace}
Ching-An Cheng, Allen Nie, and Adith Swaminathan. 2024{\natexlab{a}}.
\newblock \href {https://doi.org/10.52202/079017-2287} {{Trace} is the next {AutoDiff}: Generative optimization with rich feedback, execution traces, and {LLMs}}.
\newblock In \emph{Advances in Neural Information Processing Systems}, volume~37, pages 71596--71642.

\bibitem[{Cheng et~al.(2024{\natexlab{b}})Cheng, Cheng, He, Sun, Wang, Lin, Lian, Peng, and Hauptmann}]{cheng2024emotionllama}
Zebang Cheng, Zhi-Qi Cheng, Jun-Yan He, Jingdong Sun, Kai Wang, Yuxiang Lin, Zheng Lian, Xiaojiang Peng, and Alexander~G. Hauptmann. 2024{\natexlab{b}}.
\newblock \href {https://doi.org/10.52202/079017-3518} {{Emotion-LLaMA}: Multimodal emotion recognition and reasoning with instruction tuning}.
\newblock In \emph{Advances in Neural Information Processing Systems}, volume~37, pages 110805--110853.

\bibitem[{Eyben et~al.(2016)Eyben, Scherer, Schuller, Sundberg, Andr{\'e}, Busso, Devillers, Epps, Laukka, Narayanan, and Truong}]{eyben2016gemaps}
Florian Eyben, Klaus~R. Scherer, Bj{\"o}rn~W. Schuller, Johan Sundberg, Elisabeth Andr{\'e}, Carlos Busso, Laurence~Y. Devillers, Julien Epps, Petri Laukka, Shrikanth~S. Narayanan, and Khiet~P. Truong. 2016.
\newblock \href {https://doi.org/10.1109/TAFFC.2015.2457417} {The {Geneva} minimalistic acoustic parameter set ({GeMAPS}) for voice research and affective computing}.
\newblock \emph{IEEE Transactions on Affective Computing}, 7(2):190--202.

\bibitem[{Eyben et~al.(2010)Eyben, W{\"o}llmer, and Schuller}]{eyben2010opensmile}
Florian Eyben, Martin W{\"o}llmer, and Bj{\"o}rn Schuller. 2010.
\newblock \href {https://doi.org/10.1145/1873951.1874246} {{openSMILE}: The {Munich} versatile and fast open-source audio feature extractor}.
\newblock In \emph{Proceedings of the 18th ACM International Conference on Multimedia}, pages 1459--1462.

\bibitem[{{Gemma Team}(2026)}]{gemma2026gemma4}
{Gemma Team}. 2026.
\newblock \href {https://arxiv.org/abs/2607.02770} {{Gemma 4} technical report}.
\newblock \emph{arXiv preprint arXiv:2607.02770}.

\bibitem[{Gupta and Kembhavi(2023)}]{gupta2023visprog}
Tanmay Gupta and Aniruddha Kembhavi. 2023.
\newblock \href {https://doi.org/10.1109/CVPR52729.2023.01436} {Visual programming: Compositional visual reasoning without training}.
\newblock In \emph{Proceedings of the IEEE/CVF Conference on Computer Vision and Pattern Recognition}, pages 14953--14962.

\bibitem[{Han et~al.(2026)Han, Chan, Chen, Ai, Du, and Guo}]{han2026medconceal}
Yikun Han, Joey Chan, Jingyuan Chen, Mengting Ai, Simo Du, and Yue Guo. 2026.
\newblock \href {https://arxiv.org/abs/2604.08788} {{MedConceal}: A benchmark for clinical hidden-concern reasoning under partial observability}.
\newblock In \emph{Conference on Language Modeling (COLM)}.

\bibitem[{Holm(1979)}]{holm1979simple}
Sture Holm. 1979.
\newblock A simple sequentially rejective multiple test procedure.
\newblock \emph{Scandinavian Journal of Statistics}, 6(2):65--70.

\bibitem[{Hu et~al.(2025)Hu, Lu, and Clune}]{hu2025adas}
Shengran Hu, Cong Lu, and Jeff Clune. 2025.
\newblock \href {https://proceedings.iclr.cc/paper_files/paper/2025/hash/36b7acf6f6010652b3f2a433774a66fe-Abstract-Conference.html} {Automated design of agentic systems}.
\newblock In \emph{The Thirteenth International Conference on Learning Representations}.

\bibitem[{Huang et~al.(2024)Huang, Chen, Mishra, Zheng, Yu, Song, and Zhou}]{huang2024selfcorrect}
Jie Huang, Xinyun Chen, Swaroop Mishra, Huaixiu~Steven Zheng, Adams~Wei Yu, Xinying Song, and Denny Zhou. 2024.
\newblock \href {https://arxiv.org/abs/2310.01798} {Large language models cannot self-correct reasoning yet}.
\newblock In \emph{The Twelfth International Conference on Learning Representations}.

\bibitem[{Kapoor et~al.(2025)Kapoor, Stroebl, Siegel, Nadgir, and Narayanan}]{kapoor2025agents}
Sayash Kapoor, Benedikt Stroebl, Zachary~S. Siegel, Nitya Nadgir, and Arvind Narayanan. 2025.
\newblock \href {https://openreview.net/forum?id=Zy4uFzMviZ} {{AI} agents that matter}.
\newblock \emph{Transactions on Machine Learning Research}.

\bibitem[{Khattab et~al.(2024)Khattab, Singhvi, Maheshwari, Zhang, Santhanam, Vardhamanan, Haq, Sharma, Joshi, Moazam, Miller, Zaharia, and Potts}]{khattab2024dspy}
Omar Khattab, Arnav Singhvi, Paridhi Maheshwari, Zhiyuan Zhang, Keshav Santhanam, Sri Vardhamanan, Saiful Haq, Ashutosh Sharma, Thomas~T. Joshi, Hanna Moazam, Heather Miller, Matei Zaharia, and Christopher Potts. 2024.
\newblock \href {https://proceedings.iclr.cc/paper_files/paper/2024/hash/f1cf02ce09757f57c3b93c0db83181e0-Abstract-Conference.html} {{DSPy}: Compiling declarative language model calls into state-of-the-art pipelines}.
\newblock In \emph{The Twelfth International Conference on Learning Representations}.

\bibitem[{Kwon et~al.(2023)Kwon, Li, Zhuang, Sheng, Zheng, Yu, Gonzalez, Zhang, and Stoica}]{kwon2023vllm}
Woosuk Kwon, Zhuohan Li, Siyuan Zhuang, Ying Sheng, Lianmin Zheng, Cody~Hao Yu, Joseph~E. Gonzalez, Hao Zhang, and Ion Stoica. 2023.
\newblock \href {https://doi.org/10.1145/3600006.3613165} {Efficient memory management for large language model serving with {PagedAttention}}.
\newblock In \emph{Proceedings of the 29th Symposium on Operating Systems Principles}, pages 611--626.

\bibitem[{Lee et~al.(2026)Lee, Nair, Zhang, Lee, Khattab, and Finn}]{lee2026metaharness}
Yoonho Lee, Roshen Nair, Qizheng Zhang, Kangwook Lee, Omar Khattab, and Chelsea Finn. 2026.
\newblock \href {https://arxiv.org/abs/2603.28052} {{Meta-Harness}: End-to-end optimization of model harnesses}.
\newblock In \emph{Conference on Language Modeling (COLM)}.

\bibitem[{Li et~al.(2026{\natexlab{a}})Li, Cui, He, Wang, Zhang, Wen, and Niu}]{li2026echofoley}
Bingxuan Li, Yiming Cui, Yicheng He, Yiwei Wang, Shu Zhang, Longyin Wen, and Yulei Niu. 2026{\natexlab{a}}.
\newblock Echofoley: Event-centric hierarchical control for video grounded creative sound generation.
\newblock In \emph{Proceedings of the IEEE/CVF Conference on Computer Vision and Pattern Recognition}, pages 27229--27238.

\bibitem[{Li et~al.(2026{\natexlab{b}})Li, Hong, Qian, Ha, Liu, Wang, Guo, Li, and Ji}]{li2026trimming}
Bingxuan Li, Yining Hong, Cheng Qian, Hyeonjeong Ha, Jiateng Liu, Zhenhailong Wang, Yue Guo, Yunzhu Li, and Heng Ji. 2026{\natexlab{b}}.
\newblock Trimming the long-tail of visual world modeling evaluation.
\newblock \emph{arXiv preprint arXiv:2606.24256}.

\bibitem[{Li et~al.(2026{\natexlab{c}})Li, Kim, Qian, Chen, Anzenberg, Kundapur, and Ji}]{li-etal-2026-pearl}
Bingxuan Li, Jeonghwan Kim, Cheng Qian, Xiusi Chen, Eitan Anzenberg, Niran Kundapur, and Heng Ji. 2026{\natexlab{c}}.
\newblock \href {https://doi.org/10.18653/v1/2026.acl-long.1547} {{PEARL}: Self-evolving assistant for time management with reinforcement learning}.
\newblock In \emph{Proceedings of the 64th Annual Meeting of the {A}ssociation for {C}omputational {L}inguistics (Volume 1: Long Papers)}, pages 33522--33539, San Diego, California, United States. Association for Computational Linguistics.

\bibitem[{Li et~al.(2025)Li, Wang, Gu, Chang, and Peng}]{li2025metal}
Bingxuan Li, Yiwei Wang, Jiuxiang Gu, Kai-Wei Chang, and Nanyun Peng. 2025.
\newblock {METAL}: A multi-agent framework for chart generation with test-time scaling.
\newblock In \emph{Proceedings of the 63rd Annual Meeting of the Association for Computational Linguistics (Volume 1: Long Papers)}, pages 30054--30069.

\bibitem[{Li et~al.(2026{\natexlab{d}})Li, Shen, Liu, Xu, Liu, Li, Li, Zhu, Zhong, Lan et~al.}]{li2026toward}
Boyi Li, Yifan Shen, Yuanzhe Liu, Yifan Xu, Jiateng Liu, Xinzhuo Li, Zhengyuan Li, Jingyuan Zhu, Yunhan Zhong, Fangzhou Lan, et~al. 2026{\natexlab{d}}.
\newblock Toward cognitive supersensing in multimodal large language model.
\newblock \emph{arXiv preprint arXiv:2602.01541}.

\bibitem[{Li et~al.(2023)Li, Wei, Han, and Fan}]{li2023intentqa}
Jiapeng Li, Ping Wei, Wenjuan Han, and Lifeng Fan. 2023.
\newblock \href {https://openaccess.thecvf.com/content/ICCV2023/html/Li_IntentQA_Context-aware_Video_Intent_Reasoning_ICCV_2023_paper.html} {{IntentQA}: Context-aware video intent reasoning}.
\newblock In \emph{Proceedings of the IEEE/CVF International Conference on Computer Vision}, pages 11963--11974.

\bibitem[{Liang et~al.(2024)Liang, Goindani, Chafekar, Mathur, Yu, Salakhutdinov, and Morency}]{liang2024hemm}
Paul~Pu Liang, Akshay Goindani, Talha Chafekar, Leena Mathur, Haofei Yu, Ruslan Salakhutdinov, and Louis-Philippe Morency. 2024.
\newblock \href {https://doi.org/10.52202/079017-1358} {{HEMM}: Holistic evaluation of multimodal foundation models}.
\newblock In \emph{Advances in Neural Information Processing Systems}, volume~37, pages 42899--42940.

\bibitem[{Madaan et~al.(2023)Madaan, Tandon, Gupta, Hallinan, Gao, Wiegreffe, Alon, Dziri, Prabhumoye, Yang, Gupta, Majumder, Hermann, Welleck, Yazdanbakhsh, and Clark}]{madaan2023selfrefine}
Aman Madaan, Niket Tandon, Prakhar Gupta, Skyler Hallinan, Luyu Gao, Sarah Wiegreffe, Uri Alon, Nouha Dziri, Shrimai Prabhumoye, Yiming Yang, Shashank Gupta, Bodhisattwa~Prasad Majumder, Katherine Hermann, Sean Welleck, Amir Yazdanbakhsh, and Peter Clark. 2023.
\newblock \href {https://doi.org/10.52202/075280-2019} {{Self-Refine}: Iterative refinement with self-feedback}.
\newblock In \emph{Advances in Neural Information Processing Systems}, volume~36, pages 46534--46594.

\bibitem[{Opsahl-Ong et~al.(2024)Opsahl-Ong, Ryan, Purtell, Broman, Potts, Zaharia, and Khattab}]{opsahlong2024mipro}
Krista Opsahl-Ong, Michael~J Ryan, Josh Purtell, David Broman, Christopher Potts, Matei Zaharia, and Omar Khattab. 2024.
\newblock \href {https://doi.org/10.18653/v1/2024.emnlp-main.525} {Optimizing instructions and demonstrations for multi-stage language model programs}.
\newblock In \emph{Proceedings of the 2024 Conference on Empirical Methods in Natural Language Processing}, pages 9340--9366. Association for Computational Linguistics.

\bibitem[{Pryzant et~al.(2023)Pryzant, Iter, Li, Lee, Zhu, and Zeng}]{pryzant2023protegi}
Reid Pryzant, Dan Iter, Jerry Li, Yin~Tat Lee, Chenguang Zhu, and Michael Zeng. 2023.
\newblock \href {https://doi.org/10.18653/v1/2023.emnlp-main.494} {Automatic prompt optimization with \lq\lq gradient descent\rq\rq{} and beam search}.
\newblock In \emph{Proceedings of the 2023 Conference on Empirical Methods in Natural Language Processing}, pages 7957--7968. Association for Computational Linguistics.

\bibitem[{{Qwen Team}(2026)}]{qwen2026qwen35}
{Qwen Team}. 2026.
\newblock \href {https://qwen.ai/blog?id=qwen3.5} {{Qwen3.5}: Towards native multimodal agents}.
\newblock Blog post.

\bibitem[{Radford et~al.(2023)Radford, Kim, Xu, Brockman, McLeavey, and Sutskever}]{radford2023whisper}
Alec Radford, Jong~Wook Kim, Tao Xu, Greg Brockman, Christine McLeavey, and Ilya Sutskever. 2023.
\newblock \href {https://proceedings.mlr.press/v202/radford23a.html} {Robust speech recognition via large-scale weak supervision}.
\newblock In \emph{Proceedings of the 40th International Conference on Machine Learning}, volume 202 of \emph{Proceedings of Machine Learning Research}, pages 28492--28518. PMLR.

\bibitem[{Shen et~al.(2026)Shen, Tian, Li, Fang, Xia, Li, Jojic, Ye, Cao, Rehg et~al.}]{shen2026cogniroute}
Yifan Shen, Pei Tian, Xinzhuo Li, Bowen Fang, Shujun Xia, Bingxuan Li, Ana Jojic, Wenming Ye, Xu~Cao, James~Matthew Rehg, et~al. 2026.
\newblock Cogniroute: Learning to route social evidence in omni-modal models.
\newblock \emph{arXiv preprint arXiv:2606.20970}.

\bibitem[{Shinn et~al.(2023)Shinn, Cassano, Gopinath, Narasimhan, and Yao}]{shinn2023reflexion}
Noah Shinn, Federico Cassano, Ashwin Gopinath, Karthik Narasimhan, and Shunyu Yao. 2023.
\newblock \href {https://doi.org/10.52202/075280-0377} {{Reflexion}: Language agents with verbal reinforcement learning}.
\newblock In \emph{Advances in Neural Information Processing Systems}, volume~36, pages 8634--8652.

\bibitem[{Turpin et~al.(2023)Turpin, Michael, Perez, and Bowman}]{turpin2023unfaithful}
Miles Turpin, Julian Michael, Ethan Perez, and Samuel~R. Bowman. 2023.
\newblock \href {https://doi.org/10.52202/075280-3275} {Language models don't always say what they think: Unfaithful explanations in chain-of-thought prompting}.
\newblock In \emph{Advances in Neural Information Processing Systems}, volume~36, pages 74952--74965.

\bibitem[{Wilf et~al.(2023)Wilf, Mathur, Mathew, Ko, Kebe, Liang, and Morency}]{wilf2023socialiq}
Alex Wilf, Leena Mathur, Sheryl Mathew, Claire Ko, Youssouf Kebe, Paul~Pu Liang, and Louis-Philippe Morency. 2023.
\newblock \href {https://github.com/abwilf/Social-IQ-2.0-Challenge} {{Social-IQ 2.0} challenge: Benchmarking multimodal social understanding}.
\newblock GitHub repository.

\bibitem[{Xu et~al.(2026)Xu, Li, Li, Gu, Shen, Yu, and Wang}]{xu2026steve}
Yifan Xu, Yixuan Li, Xinzhuo Li, Yixin Gu, Yifan Shen, Lijun Yu, and Haohan Wang. 2026.
\newblock Steve: Stabilizing textual gradient-based prompt optimization via error-driven refinement and regularized verification.
\newblock \emph{arXiv preprint arXiv:2609.23716}.

\bibitem[{Yang et~al.(2024)Yang, Wang, Lu, Liu, Le, Zhou, and Chen}]{yang2024opro}
Chengrun Yang, Xuezhi Wang, Yifeng Lu, Hanxiao Liu, Quoc~V. Le, Denny Zhou, and Xinyun Chen. 2024.
\newblock \href {https://proceedings.iclr.cc/paper_files/paper/2024/hash/3339f19c5fcee3ad74502947a32be9e6-Abstract-Conference.html} {Large language models as optimizers}.
\newblock In \emph{The Twelfth International Conference on Learning Representations}.

\bibitem[{Yang et~al.(2026)Yang, Zhao, Wu, and K{\"a}stner}]{yang2026better}
Chenyang Yang, Xinran Zhao, Tongshuang Wu, and Christian K{\"a}stner. 2026.
\newblock \href {https://arxiv.org/abs/2607.08938} {Better harnesses, smaller models: Building 90\% cheaper agents via automated harness adaptation}.
\newblock \emph{arXiv preprint arXiv:2607.08938}.

\bibitem[{Yuksekgonul et~al.(2025)Yuksekgonul, Bianchi, Boen, Liu, Lu, Huang, Guestrin, and Zou}]{yuksekgonul2025textgrad}
Mert Yuksekgonul, Federico Bianchi, Joseph Boen, Sheng Liu, Pan Lu, Zhi Huang, Carlos Guestrin, and James Zou. 2025.
\newblock \href {https://doi.org/10.1038/s41586-025-08661-4} {Optimizing generative {AI} by backpropagating language model feedback}.
\newblock \emph{Nature}, 639(8055):609--616.

\bibitem[{Zhang et~al.(2026)Zhang, Cheng, Deng, Li, Lian, Chen, Liu, Wang, Zhang, Zhang, Guo, Zhu, Wu, Wang, Zheng, Peng, Wu, Wang, Li, Ye, and Heng}]{zhang2025mme}
Fan Zhang, Zebang Cheng, Chong Deng, Haoxuan Li, Zheng Lian, Qian Chen, Huadai Liu, Wen Wang, Yi-Fan Zhang, Renrui Zhang, Ziyu Guo, Zhihong Zhu, Hao Wu, Haixin Wang, Yefeng Zheng, Xiaojiang Peng, Xian Wu, Kun Wang, Xiangang Li, and 2 others. 2026.
\newblock \href {https://proceedings.iclr.cc/paper_files/paper/2026/hash/50d277e84b2bcbaadcd84548a87e8cc4-Abstract-Conference.html} {{MME-Emotion}: A holistic evaluation benchmark for emotional intelligence in multimodal large language models}.
\newblock In \emph{The Fourteenth International Conference on Learning Representations}.

\bibitem[{Zhang et~al.(2025)Zhang, Xiang, Yu, Teng, Chen, Chen, Zhuge, Cheng, Hong, Wang, Zheng, Liu, Luo, and Wu}]{zhang2025aflow}
Jiayi Zhang, Jinyu Xiang, Zhaoyang Yu, Fengwei Teng, Xionghui Chen, Jiaqi Chen, Mingchen Zhuge, Xin Cheng, Sirui Hong, Jinlin Wang, Bingnan Zheng, Bang Liu, Yuyu Luo, and Chenglin Wu. 2025.
\newblock \href {https://proceedings.iclr.cc/paper_files/paper/2025/hash/5492ecbce4439401798dcd2c90be94cd-Abstract-Conference.html} {{AFlow}: Automating agentic workflow generation}.
\newblock In \emph{The Thirteenth International Conference on Learning Representations}.

\bibitem[{Zhao et~al.(2025)Zhao, Wei, and Bo}]{zhao2025r1omni}
Jiaxing Zhao, Xihan Wei, and Liefeng Bo. 2025.
\newblock \href {https://arxiv.org/abs/2503.05379} {{R1-Omni}: Explainable omni-multimodal emotion recognition with reinforcement learning}.
\newblock \emph{arXiv preprint arXiv:2503.05379}.

\bibitem[{Zheng et~al.(2023)Zheng, Chiang, Sheng, Zhuang, Wu, Zhuang, Lin, Li, Li, Xing, Zhang, Gonzalez, and Stoica}]{zheng2023judging}
Lianmin Zheng, Wei-Lin Chiang, Ying Sheng, Siyuan Zhuang, Zhanghao Wu, Yonghao Zhuang, Zi~Lin, Zhuohan Li, Dacheng Li, Eric~P. Xing, Hao Zhang, Joseph~E. Gonzalez, and Ion Stoica. 2023.
\newblock \href {https://doi.org/10.52202/075280-2020} {Judging {LLM}-as-a-judge with {MT-Bench} and {Chatbot Arena}}.
\newblock In \emph{Advances in Neural Information Processing Systems}, volume~36, pages 46595--46623.

\bibitem[{Zhou et~al.(2023)Zhou, Muresanu, Han, Paster, Pitis, Chan, and Ba}]{zhou2023ape}
Yongchao Zhou, Andrei~Ioan Muresanu, Ziwen Han, Keiran Paster, Silviu Pitis, Harris Chan, and Jimmy Ba. 2023.
\newblock \href {https://openreview.net/forum?id=92gvk82DE-} {Large language models are human-level prompt engineers}.
\newblock In \emph{The Eleventh International Conference on Learning Representations}.

\bibitem[{Zhu et~al.(2025)Zhu, Liu, Li, Tian, Yang, Zhang, Han, Xie, Cui, Zhang et~al.}]{zhu2025llm}
Kunlun Zhu, Zijia Liu, Bingxuan Li, Muxin Tian, Yingxuan Yang, Jiaxun Zhang, Pengrui Han, Qipeng Xie, Fuyang Cui, Weijia Zhang, et~al. 2025.
\newblock \href {https://arxiv.org/abs/2509.25370} {Where {LLM} agents fail and how they can learn from failures}.
\newblock \emph{arXiv preprint arXiv:2509.25370}.

\end{thebibliography}
\clearpage
\appendix
\section{Implementation Details}
\label{app:implementation}
\label{app:provenance}
\paragraph{Scoring and statistics.}
All recognition means and QA accuracies are computed from item-level predictions. \mheshort{}$-$Shared and downstream intervals use paired bootstraps (task-stratified for recognition, clustered by video for QA); Audio$-$Video, Shared$-$Audio, and per-condition intervals use the same task-stratified procedure. All bootstraps draw 10,000 resamples.

\paragraph{Evaluation split.}
The evaluation set (5,507 items on 5,083 media) excludes 70 items whose media overlap construction, validation, or the 129 reserved examples. It contains 986 items that were inspected during development and 4,521 that were not; because we cannot identify which items were inspected, results excluding them cannot be reported. Inspection could have influenced any human design choice, most plausibly the Shared scaffold. It did not inform the Builder's edits directly, because the Builder never saw evaluation items.

\begin{figure}[t]
\centering
\includegraphics[width=\columnwidth]{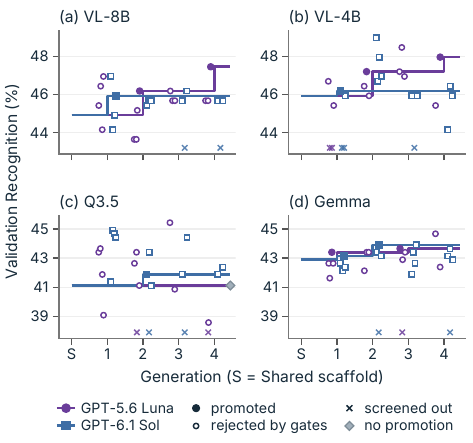}
\caption{\textbf{\mheshort{} trajectories on the validation set.} Incumbent validation Recognition (item mean over 394 items) per generation for each Builder; markers are candidates scored on validation (filled: promoted; open: rejected by the gates); $\times$ marks proposals screened out by the checks or the reviewer, drawn at the panel floor and never scored. S: Shared scaffold; diamond: run with no promotion.}
\label{fig:trajectories}
\end{figure}

\begin{figure}[t]
\centering
\includegraphics[width=\columnwidth]{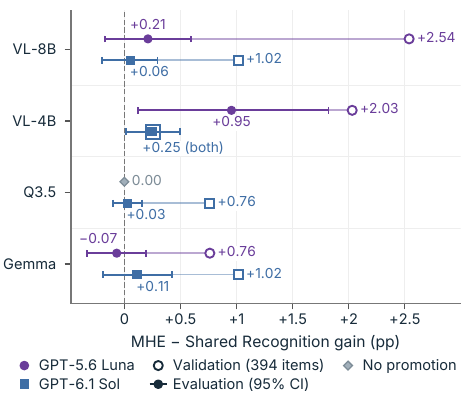}
\caption{\textbf{Validation gains overstate evaluation gains.} Gain of each evolved harness over Shared on validation (item mean over 394 items) and on evaluation (task macro over 5,507 items). Connected points describe the same harness under two summaries, not two estimates of one quantity.}
\label{fig:construction_transfer}
\end{figure}

\paragraph{Construction runs.}
Each harness comes from one four-generation run. The GPT-6.1 Sol runs were made about three weeks after the GPT-5.6 Luna runs, and their diagnostic packet lists one additional tool (about 58 tokens); first-generation inputs are otherwise identical across Builders. Each run's validation trajectory follows the incumbent after each generation's promotion decision. Table~\ref{tab:builds} lists the construction records, Figure~\ref{fig:trajectories} traces every generation, and Figure~\ref{fig:construction_transfer} contrasts validation and evaluation gains. A third Builder configuration, also run on all four students, is not analyzed further: one of its 24 validation-scored proposals was promoted (for Qwen3-VL-4B, worth $+2.03$ validation points but $+0.04$ task-macro points on evaluation), and including it leaves every conclusion unchanged (12-test families; smallest adjusted $p$: \mheshort{}$-$Shared .288, IntentQA .127).

\begin{table}[t]
\centering
\footnotesize
\setlength{\tabcolsep}{2.4pt}
\renewcommand{\arraystretch}{1.1}
\begin{tabularx}{\columnwidth}{@{}cl*{6}{>{\centering\arraybackslash}X}@{}}
\toprule
 & & & & \multicolumn{2}{c}{$\mathcal{D}_v$ Rec.\ (\%)} & Seen & \\
\cmidrule(lr){5-6}
 & Student & Scored & Prom. & Start & Final & eval. & USD \\
\midrule
 & VL-8B & 14 & 2 & 44.92 & 47.46 & $+0.10$ & 0.41 \\
 & VL-4B & 10 & 2 & 45.94 & 47.97 & $+0.20$ & 0.49 \\
 & Q3.5 & 10 & 0 & 41.12 & 41.12 & $0.00$\rlap{$^{\dagger}$} & 0.23 \\
\multirow{-4}{*}{\rotatebox[origin=c]{90}{\scriptsize GPT-5.6 Luna}} & Gemma & 12 & 2 & 42.89 & 43.65 & $+0.10$ & 0.40 \\
\midrule
 & VL-8B & 12 & 1 & 44.92 & 45.94 & $+0.30$ & 1.51 \\
 & VL-4B & 11 & 1 & 45.94 & 46.19 & $+0.30$ & 1.49 \\
 & Q3.5 & 11 & 1 & 41.12 & 41.88 & $+0.10$ & 1.53 \\
\multirow{-4}{*}{\rotatebox[origin=c]{90}{\scriptsize GPT-6.1 Sol}} & Gemma & 14 & 2 & 42.89 & 43.91 & $-0.20$ & 1.65 \\
\bottomrule
\end{tabularx}
\captionsetup{position=bottom}
\caption{\textbf{\mheshort{} construction records.} Validation Recognition is a flat item mean over 394 items, distinct from evaluation task-macro Recognition. Every run has four generations. Scored: proposals scored on $\mathcal{D}_v$; Prom.: promotions. Proposals screened out by static checks or the proposal reviewer before validation: 5 (GPT-5.6 Luna) and 9 (GPT-6.1 Sol). USD covers logged Builder, reviewer, and proxy-judge calls, excluding local GPU work. Seen eval.: gain over Shared as a flat item mean on the 2{,}963 seen-type evaluation items, comparable to Final$-$Start. VL-8B/VL-4B: Qwen3-VL-8B/4B; Q3.5: Qwen3.5-4B; Gemma: Gemma-4-E4B. $^{\dagger}$\,Retained scaffold with identical labels.}
\label{tab:builds}
\end{table}

\paragraph{Serving and editable interfaces.}
Native video is decoded at 2~fps with at most 32 frames; auxiliary frame-cache sampling is a separate path, and the Shared scaffold attaches no auxiliary images or face crops to later stages. Emotion-context generation uses a 16,384-token context window and downstream answering uses 32,768. Prompt and configuration edits target existing prompt and YAML files, and full-bundle search also permits restricted verification Python files; perception tools, schemas, and the output controller are outside that scope.

\section{Execution and Construction Details}
\label{app:method}
This appendix expands on the execution and construction rules in Section~\ref{sec:inference}; it defines no additional evaluated conditions.

\begin{figure}[t]
\centering
\fbox{\begin{minipage}{0.91\columnwidth}
\small
\textbf{\mhe{}}\\[3pt]
\textbf{Input:} student $M_\theta$, Shared scaffold $H_0$, construction data $\mathcal{D}_b$, validation data $\mathcal{D}_v$, generations $T$.\\[2pt]
$H\leftarrow H_0$; initialize the proposal history and the solved set $\mathcal{S}$ from $H$.\\
For $t=1,\ldots,T$:
\begin{enumerate}
\setlength{\itemsep}{1pt}
\setlength{\parskip}{0pt}
\item Run $H$ on $\mathcal{D}_b$; cluster failures by gold--prediction confusion and cross-modal conflict.
\item \emph{Variation:} the Builder proposes up to four candidate harnesses, editing prompts, settings, and restricted verification code.
\item Reject candidates that fail the static checks or the proposal reviewer.
\item \emph{Selection:} score survivors on $\mathcal{D}_v$ (computing the reasoning proxy only for those passing the cheaper gates) and apply Eq.~\ref{eq:gates}.
\item Promote the best passing candidate or retain $H$; update $\mathcal{S}$ and the history.
\end{enumerate}
\textbf{Output:} the frozen evolved harness $H^\star=H$, evaluated only after construction.
\end{minipage}}
\refstepcounter{algorithm}\label{alg:builder}\par
\smallskip\begin{minipage}{\columnwidth}\fontsize{10pt}{12pt}\selectfont Algorithm~\thealgorithm:\ \textbf{\mhe{}.} The Builder proposes edits to the harness, and checks and validation gates decide which survive. The Builder never answers evaluation items.\end{minipage}
\end{figure}

\subsection{Perception and Evidence Assembly}
Whisper produces word-timestamped transcripts. Acoustic features are verbalized relative to reference means and standard deviations: an absolute standardized deviation of at least one yields a descriptor. Missing features and zero-variance reference features are omitted, and the descriptions neither identify speakers nor label emotion.

Native-video decoding and auxiliary frame-cache sampling are separate paths (Appendix~\ref{app:implementation}). Deterministic assembly assigns identifiers $\texttt{v1},\ldots$ to visual observations and $\texttt{a1},\ldots$ to transcript sentences and prosody phrases. The request target is stored as \texttt{t1} but cannot be cited as observed decision evidence. Each hypothesis has a label, a model-generated confidence, supporting identifiers, and contradicting identifiers. Valid identifiers do not establish semantic support.

\subsection{Output and Failure Handling}
The controller enforces the requested vocabulary and cardinality, with a three-label cap for multi-label outputs. A candidate normally needs a valid supporting identifier and more support than contradiction. If too few pass, remaining slots are filled from available rankings and labels. This is completion logic, not an evidence guarantee.

The \emph{Explain} call receives the record and verified interpretations and writes context, facial and bodily cues, vocal and verbal cues, and a cross-modal synthesis with the strongest alternative. The controller sanitizes citations and appends a decision section. With upstream outputs fixed, the explanation cannot change the labels, but it can change the downstream context. One retry per malformed stage is permitted under the shared six-call cap; once the cap is reached, the remaining stages are skipped and the output is completed from available information.

\subsection{Edit Spaces and Selection History}
All runs search the full edit space (existing prompts, YAML settings, and restricted verification files) directly from Shared. Each proposal supplies complete replacements for the allowed files and names its target construction-failure cluster. The solved set $\mathcal{S}_t$ (Eq.~\ref{eq:regression}) grows with each promotion. The proposal history contains proposals, construction effects, and check and reviewer feedback; validation scores and gate decisions are withheld, although validation still decides which incumbent supplies the next diagnostics.

\mheshort{} obtains reasoning-proxy scores for the initial harness and for candidates that pass the cheaper recognition, regression, and call-budget gates before applying Eq.~\ref{eq:gates}. Static checks and the proposal reviewer constrain candidates but cannot guarantee their safety or semantic correctness.

\paragraph{Reasoning proxy used for selection.}
We evaluate $\widehat Q_v$ on a fixed 200-item sample from $\mathcal{D}_v$. The judge receives each response, up to ten cached frames, and gold labels; the benchmark's judge-side audio-clue field is left empty. Its rubric gives binary scores to non-final reasoning steps, marking clear contradictions with the supplied cues as incorrect. The per-item proxy is the mean of these non-final scores, or zero if none exist; $\widehat Q_v$ averages the per-item values. The final decision-step score is excluded. Following the LLM-as-judge protocol of \citet{zheng2023judging}, the judge is GPT-5.6 Luna. The proxy does not measure official reasoning performance or explanation faithfulness.

\section{Detailed Recognition Results}
\label{app:recognition}
\begin{figure}[t]
\centering
\includegraphics[width=\columnwidth]{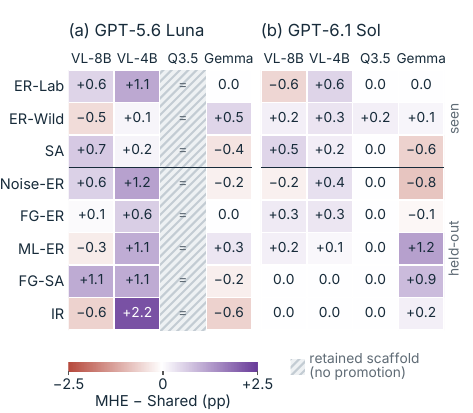}
\caption{\textbf{Per-task \mheshort{}$-$Shared changes.} Differences (pp) for each Builder. Both panels share one zero-centered color scale, and the horizontal rule separates the three seen task types from the five held-out types. Exact scores are in Table~\ref{tab:main}.}
\label{fig:task_adaptation}
\end{figure}

\begin{figure}[t]
\centering
\includegraphics[width=\columnwidth]{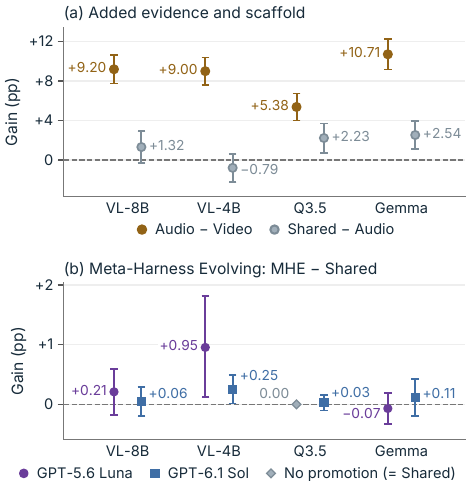}
\caption{\textbf{Decomposing recognition gains.} Task-macro contrasts (pp) on 5,507 MME-Emotion items. \textbf{(a)}~Audio-derived text adds 5.4--10.7 points, and the Shared scaffold changes Audio by between $-0.8$ and $+2.5$. \textbf{(b)}~\mheshort{} adds at most $+0.95$ over Shared (note the narrower axis), and no contrast survives Holm correction over eight tests. Whiskers are pointwise 95\% paired bootstrap intervals; the diamond marks the run that retained the scaffold.}
\label{fig:recognition_gains}
\end{figure}

Figure~\ref{fig:recognition_gains} and Table~\ref{tab:builder_inference} report the Audio$-$Video and Shared$-$Audio contrasts and the eight \mheshort{}$-$Shared contrasts, the latter with Holm-adjusted probabilities; a pointwise interval that excludes zero can coexist with a nonsignificant corrected test. Per-task Recognition for all five conditions appears in Table~\ref{tab:main}, and Figure~\ref{fig:task_adaptation} visualizes the per-task \mheshort{}$-$Shared changes. Task-level values are descriptive and not separately corrected.

\begin{table}[t]
\centering
\footnotesize
\setlength{\tabcolsep}{4pt}
\renewcommand{\arraystretch}{1.1}
{\raggedright\textit{(a) \mhe{}: \mheshort{}$-$Shared}\par}\smallskip
\begin{tabularx}{\columnwidth}{@{}l>{\raggedright\arraybackslash}Xrrr@{}}
\toprule
 & Builder & $\Delta$ & 95\% CI & $p_{\mathrm{Holm}}$ \\
\midrule
 & \mbox{GPT-5.6 Luna} & $+0.21$ & \textcolor{inkGray}{\scriptsize $-0.18$ to $+0.59$} & 1.000 \\
\multirow{-2}{*}{VL-8B} & \mbox{GPT-6.1 Sol} & $+0.06$ & \textcolor{inkGray}{\scriptsize $-0.20$ to $+0.29$} & 1.000 \\
\midrule
 & \mbox{GPT-5.6 Luna} & $+0.95$ & \textcolor{inkGray}{\scriptsize $+0.12$ to $+1.82$} & .192 \\
\multirow{-2}{*}{VL-4B} & \mbox{GPT-6.1 Sol} & $+0.25$ & \textcolor{inkGray}{\scriptsize $+0.01$ to $+0.49$} & .262 \\
\midrule
 & \mbox{GPT-5.6 Luna} & $0.00$\rlap{$^{\dagger}$} & \textcolor{inkGray}{\scriptsize $0.00$ to $0.00$} & 1.000 \\
\multirow{-2}{*}{Q3.5} & \mbox{GPT-6.1 Sol} & $+0.03$ & \textcolor{inkGray}{\scriptsize $-0.10$ to $+0.16$} & 1.000 \\
\midrule
 & \mbox{GPT-5.6 Luna} & $-0.07$ & \textcolor{inkGray}{\scriptsize $-0.33$ to $+0.20$} & 1.000 \\
\multirow{-2}{*}{Gemma} & \mbox{GPT-6.1 Sol} & $+0.11$ & \textcolor{inkGray}{\scriptsize $-0.19$ to $+0.42$} & 1.000 \\
\bottomrule
\end{tabularx}
\par\medskip
{\raggedright\textit{(b) Added evidence and scaffold}\par}\smallskip
\begin{tabularx}{\columnwidth}{@{}l>{\raggedright\arraybackslash}Xrr@{}}
\toprule
 & Contrast & $\Delta$ & 95\% CI \\
\midrule
 & Audio$-$Video & $+9.20$ & \textcolor{inkGray}{\scriptsize $+7.76$ to $+10.63$} \\
\multirow{-2}{*}{VL-8B} & Shared$-$Audio & $+1.32$ & \textcolor{inkGray}{\scriptsize $-0.28$ to $+2.92$} \\
\midrule
 & Audio$-$Video & $+9.00$ & \textcolor{inkGray}{\scriptsize $+7.63$ to $+10.39$} \\
\multirow{-2}{*}{VL-4B} & Shared$-$Audio & $-0.79$ & \textcolor{inkGray}{\scriptsize $-2.25$ to $+0.61$} \\
\midrule
 & Audio$-$Video & $+5.38$ & \textcolor{inkGray}{\scriptsize $+4.02$ to $+6.76$} \\
\multirow{-2}{*}{Q3.5} & Shared$-$Audio & $+2.23$ & \textcolor{inkGray}{\scriptsize $+0.73$ to $+3.70$} \\
\midrule
 & Audio$-$Video & $+10.71$ & \textcolor{inkGray}{\scriptsize $+9.16$ to $+12.29$} \\
\multirow{-2}{*}{Gemma} & Shared$-$Audio & $+2.54$ & \textcolor{inkGray}{\scriptsize $+1.12$ to $+3.95$} \\
\bottomrule
\end{tabularx}
\captionsetup{position=bottom}
\caption{\textbf{Recognition contrasts with intervals.} Task-macro Recognition differences (pp) with pointwise 95\% intervals. (a) \mheshort{}$-$Shared, with task-stratified paired bootstrap intervals (10{,}000 resamples) and Holm-adjusted $p$. No difference survives Holm correction over eight tests (smallest adjusted $p=.192$). $^{\dagger}$\,Retained scaffold with identical labels. (b) Audio$-$Video and Shared$-$Audio, with intervals from the same procedure (pointwise; not part of a Holm family). VL-8B/VL-4B: Qwen3-VL-8B/4B; Q3.5: Qwen3.5-4B; Gemma: Gemma-4-E4B.}
\label{tab:builder_inference}
\end{table}

\paragraph{Metric interpretation.}
ER-Lab, ER-Wild, and SA are the seen types (177, 1,491, and 1,295 items); Noise-ER, FG-ER, ML-ER, FG-SA, and IR are held out (500, 500, 590, 461, and 493 items). FG-ER and ML-ER use recall-like multi-label Recognition, which should not be read as precision, exact match, or balanced F1. This choice favors the scaffold: on these two tasks Shared predicts 2.2--2.9 labels per item against 1.5--2.0 for Audio. Scoring FG-ER and ML-ER by set F1 instead changes Shared$-$Audio to $-0.25$, $-1.41$, $+1.10$, and $+1.00$ points (VL-8B, VL-4B, Q3.5, Gemma), while Audio$-$Video ($+6.05$ to $+11.17$) and \mheshort{}$-$Shared (at most $+0.80$) barely move. Under set F1, every evolved harness for the two Qwen3-VL students scores below single-call Audio ($-0.05$ to $-1.16$ points), so for them the four-call harness adds nothing over the audio-derived evidence.

\section{Resource Accounting}
\label{app:resources}
Table~\ref{tab:builds} reports the API charges logged for each construction run, including prompt-cache discounts. They cover the Builder, the proposal reviewer, and the proxy judge, and exclude local GPU work and fixed preprocessing, so they are not a matched end-to-end efficiency comparison. Each construction run cost \$0.23--0.49 with GPT-5.6 Luna and \$1.49--1.65 with GPT-6.1 Sol. All student inference ran locally with vLLM on two NVIDIA A100 80GB GPUs; we did not record GPU-hours, so we report call volume instead. The eight construction runs issued 9,824--25,312 student calls each (162,007 in total; 292M input and 30M output tokens), MME-Emotion evaluation issued 309,226 student calls across the five reported conditions, and the downstream study generated 11,848 emotion contexts and answered 78,448 questions. Inference with Shared or an evolved harness uses about four student calls per item instead of one. For $V$ videos and $Q$ questions, reusing emotion context costs
\begin{equation}
\begin{aligned}
C_{\mathrm{reuse}}={}&C_{\mathrm{build}}+
V(C_{\mathrm{perception}}+C_{\mathrm{emotion}})\\
&{}+Q\, C_{\mathrm{QA+context}}.
\end{aligned}
\end{equation}
Reuse amortizes context generation across questions but repeats context tokens in every answer; this accounting does not establish a cost advantage.

\section{Downstream Protocol and Diagnostics}
\label{app:downstream}
The eight primary Audio+Emotion$-$Audio contrasts in Table~\ref{tab:context_delta} have pointwise intervals and probabilities Holm-adjusted over the eight tests; none survives correction. The context contains the predicted emotion, its confidence, alternatives, and analyses of setting, facial and bodily cues, vocal and verbal cues, and cross-modal interpretation; it is generated without any downstream question.

\begin{table}[t]
\centering
\footnotesize
\setlength{\tabcolsep}{3.5pt}
\renewcommand{\arraystretch}{1.1}
\begin{tabularx}{\columnwidth}{@{}l>{\raggedright\arraybackslash}Xrrr@{}}
\toprule
Student & Builder & $\Delta$ & 95\% CI & $p_{\mathrm{Holm}}$ \\
\midrule
VL-8B & \mbox{GPT-5.6 Luna} & $+0.47$ & \textcolor{inkGray}{\scriptsize $-1.15$ to $+2.05$} & 1.000 \\
 & \mbox{GPT-6.1 Sol} & $+0.23$ & \textcolor{inkGray}{\scriptsize $-1.33$ to $+1.77$} & 1.000 \\
VL-4B & \mbox{GPT-5.6 Luna} & $+1.87$ & \textcolor{inkGray}{\scriptsize $+0.47$ to $+3.27$} & .085 \\
 & \mbox{GPT-6.1 Sol} & $+1.50$ & \textcolor{inkGray}{\scriptsize $+0.05$ to $+2.97$} & .344 \\
Q3.5 & \mbox{GPT-5.6 Luna} & $-1.50$\rlap{$^{\dagger}$} & \textcolor{inkGray}{\scriptsize $-3.05$ to $+0.09$} & .395 \\
 & \mbox{GPT-6.1 Sol} & $-1.41$ & \textcolor{inkGray}{\scriptsize $-2.91$ to $+0.15$} & .426 \\
Gemma & \mbox{GPT-5.6 Luna} & $+1.03$ & \textcolor{inkGray}{\scriptsize $-0.63$ to $+2.66$} & .695 \\
 & \mbox{GPT-6.1 Sol} & $+1.36$ & \textcolor{inkGray}{\scriptsize $-0.24$ to $+2.96$} & .426 \\
\bottomrule
\end{tabularx}
\captionsetup{position=bottom}
\caption{\textbf{Effect of emotion context on downstream QA.} Audio+Emotion$-$Audio accuracy (pp) on the 2{,}134 IntentQA test questions, with pointwise 95\% intervals from a paired bootstrap clustered by video (10{,}000 resamples). No contrast survives Holm correction over eight tests (smallest adjusted $p=.085$). $^{\dagger}$\,Retained scaffold: this context is generated by Shared.}
\label{tab:context_delta}
\end{table}

\begin{figure}[t]
\centering
\includegraphics[width=\columnwidth]{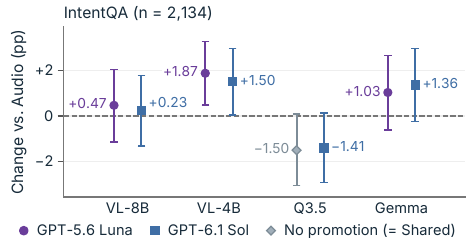}
\caption{\textbf{Emotion context added to Audio on IntentQA.} Accuracy change (pp); whiskers: pointwise 95\% paired intervals clustered by video. No contrast survives Holm correction (eight tests). Frames and audio text are identical in both conditions; the gray diamond (Qwen3.5-4B, GPT-5.6 Luna) is the retained Shared scaffold's context.}
\label{fig:context}
\end{figure}

\begin{table*}[t]
\centering
\footnotesize
\setlength{\tabcolsep}{2pt}
\renewcommand{\arraystretch}{1.1}
\begin{tabularx}{\textwidth}{@{}l*{8}{>{\centering\arraybackslash}X}@{}}
\toprule
 & \multicolumn{2}{c}{VL-8B} & \multicolumn{2}{c}{VL-4B} & \multicolumn{2}{c}{Q3.5} & \multicolumn{2}{c}{Gemma} \\
\cmidrule(lr){2-3}\cmidrule(lr){4-5}\cmidrule(lr){6-7}\cmidrule(lr){8-9}
Stratum & Luna & Sol & Luna & Sol & Luna & Sol & Luna & Sol \\
\midrule
Why (CW) & \mbox{$+1.04$} & \mbox{$+0.48$} & \mbox{$+1.36$} & \mbox{$+1.04$} & \mbox{$-1.68$} & \mbox{$-1.76$} & \mbox{$+0.96$} & \mbox{$+1.44$} \\
{\scriptsize $n=1{,}250$} & \mbox{\textcolor{inkGray}{\scriptsize $-1.0$ to $+3.1$}} & \mbox{\textcolor{inkGray}{\scriptsize $-1.5$ to $+2.4$}} & \mbox{\textcolor{inkGray}{\scriptsize $-0.6$ to $+3.3$}} & \mbox{\textcolor{inkGray}{\scriptsize $-0.9$ to $+2.9$}} & \mbox{\textcolor{inkGray}{\scriptsize $-3.6$ to $+0.2$}} & \mbox{\textcolor{inkGray}{\scriptsize $-3.6$ to $+0.2$}} & \mbox{\textcolor{inkGray}{\scriptsize $-1.1$ to $+3.2$}} & \mbox{\textcolor{inkGray}{\scriptsize $-0.6$ to $+3.6$}} \\
How (CH) & \mbox{$-2.23$} & \mbox{$-2.51$} & \mbox{$+2.51$} & \mbox{$+3.90$} & \mbox{$-0.84$} & \mbox{$-0.56$} & \mbox{$+0.84$} & \mbox{$+1.11$} \\
{\scriptsize $n=359$} & \mbox{\textcolor{inkGray}{\scriptsize $-6.0$ to $+1.4$}} & \mbox{\textcolor{inkGray}{\scriptsize $-6.2$ to $+1.2$}} & \mbox{\textcolor{inkGray}{\scriptsize $-0.6$ to $+5.7$}} & \mbox{\textcolor{inkGray}{\scriptsize $+0.3$ to $+7.4$}} & \mbox{\textcolor{inkGray}{\scriptsize $-4.6$ to $+2.8$}} & \mbox{\textcolor{inkGray}{\scriptsize $-4.3$ to $+3.1$}} & \mbox{\textcolor{inkGray}{\scriptsize $-2.9$ to $+4.6$}} & \mbox{\textcolor{inkGray}{\scriptsize $-2.6$ to $+4.9$}} \\
Temporal & \mbox{$+0.95$} & \mbox{$+1.52$} & \mbox{$+2.67$} & \mbox{$+0.95$} & \mbox{$-1.52$} & \mbox{$-1.14$} & \mbox{$+1.33$} & \mbox{$+1.33$} \\
{\scriptsize $n=525$} & \mbox{\textcolor{inkGray}{\scriptsize $-2.4$ to $+4.3$}} & \mbox{\textcolor{inkGray}{\scriptsize $-1.7$ to $+4.9$}} & \mbox{\textcolor{inkGray}{\scriptsize $-0.4$ to $+5.8$}} & \mbox{\textcolor{inkGray}{\scriptsize $-2.2$ to $+4.2$}} & \mbox{\textcolor{inkGray}{\scriptsize $-5.0$ to $+1.8$}} & \mbox{\textcolor{inkGray}{\scriptsize $-4.7$ to $+2.2$}} & \mbox{\textcolor{inkGray}{\scriptsize $-2.0$ to $+4.7$}} & \mbox{\textcolor{inkGray}{\scriptsize $-2.1$ to $+4.7$}} \\
\bottomrule
\end{tabularx}
\captionsetup{position=bottom}
\caption{\textbf{Emotion context by IntentQA question type.} Audio+Emotion$-$Audio accuracy (pp) within exploratory strata given by the official types of the 2{,}134 test questions, never by outcomes: causal-why (CW), causal-how (CH), and temporal (TN+TP, next and previous); every stratum is reported. Gray: pointwise 95\% interval from a paired bootstrap clustered by video (10{,}000 resamples); $n$: questions in the stratum. No contrast survives Holm correction within its stratum (8 tests: 4 students $\times$ 2 Builders; smallest adjusted $p=.278$). Luna: GPT-5.6~Luna; Sol: GPT-6.1~Sol. VL-8B/VL-4B: Qwen3-VL-8B/4B; Q3.5: Qwen3.5-4B; Gemma: Gemma-4-E4B.}
\label{tab:downstream_strata}
\end{table*}
\section{Additional Case Studies}
\label{app:cases}
\begin{figure*}[t]
\centering
\includegraphics[width=\textwidth,height=0.77\textheight,keepaspectratio]{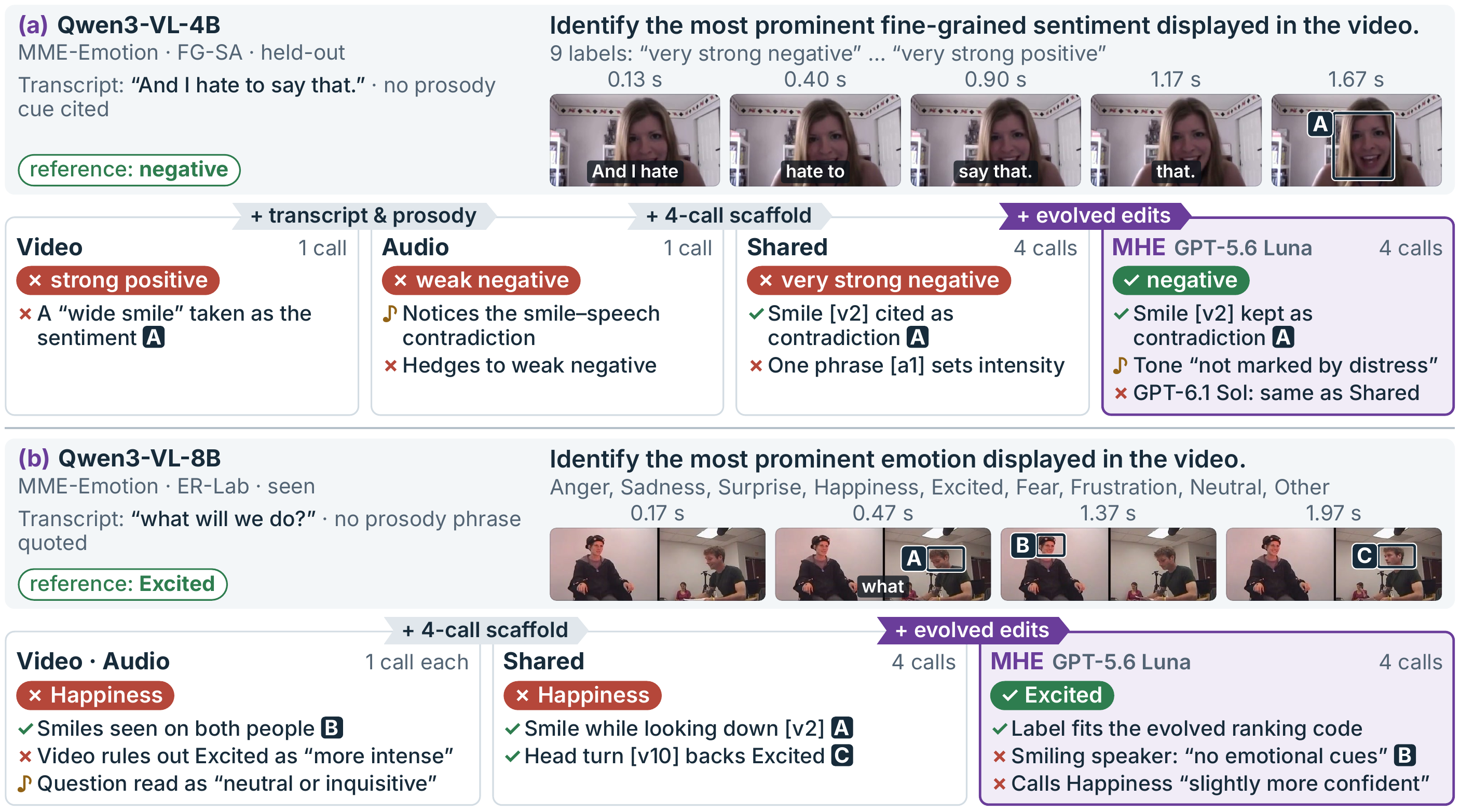}
\caption{\textbf{Additional cases.} Selected, not representative; letters, note glyphs, and subtitles as in Figure~\ref{fig:case} (no headline tags or edit panels); merged cards report two conditions that agree. \textbf{(a)}~Qwen3-VL-4B, FG-SA: the GPT-5.6 Luna harness reaches the reference \emph{negative}; GPT-6.1 Sol's does not. \textbf{(b)}~Qwen3-VL-8B, ER-Lab: the correct \emph{Excited} fits the evolved ranking code, while the written explanation favors \emph{Happiness}.}
\label{fig:case_appendix}
\end{figure*}
Figure~\ref{fig:case_appendix} shows two further selected MME-Emotion cases in the layout of Figure~\ref{fig:case}. In (a), for Qwen3-VL-4B, a clip pairs a smile with the words ``And I hate to say that''; the GPT-5.6 Luna harness reaches the reference \emph{negative}, whereas the GPT-6.1 Sol harness does not. In (b), the Qwen3-VL-8B harness evolved by GPT-5.6 Luna switches Shared's \emph{Happiness} to the correct \emph{Excited}, while its written explanation still leans toward \emph{Happiness}. The switch is consistent with its evolved ranking code, which boosts audio-supported alternatives, although its intermediate ranking is unavailable; explanations need not track the decision rule.

\section{Controls for Future Work}
\label{app:pending}
The controls below would sharpen the comparisons in this paper.

\paragraph{Evidence and computation.}
Transcript-only and prosody-only conditions would separate the sources of the audio gain. A one-call structured prediction using the same tools and controller would test decomposition against a direct call that uses the same output contract. Repeated direct samples with a fixed aggregation rule would test the value of additional calls. Token and call counts must be measured rather than presumed matched. Giving the decision stages access to the media would measure how much information is lost when observations are converted to text.

\paragraph{Construction and context.}
Comparing prompt- or configuration-only search with full-bundle search requires matched splits and budgets. Evaluating the twelve candidates that the regression gate rejected despite gains of at least eight validation items, or rerunning selection with a looser $\delta$, would test whether the gate rather than the proposals limits \mheshort{}. Independent construction repetitions are necessary to assess search variability. Shared-generated contexts for every student are needed to test the downstream increment from evolution. Length-matched visual captions and non-emotion summaries would separate emotion-specific utility from generic additional text. The reserved examples require a separately frozen evaluation, and explanation quality requires its own measurement.

\end{document}